# A learning perspective on the emergence of abstractions: The curious case of phone(me)s

Petar MILIN[*]

*University of Birmingham, Department of Modern Languages, Birmingham, UK*

Benjamin V. TUCKER

*University of Alberta, Department of Linguistics, Edmonton, Alberta, Canada*

Dagmar DIVJAK

*University of Birmingham, Department of Modern Languages & Department of English Language and Linguistics, Birmingham, UK*

In the present paper we use a range of modeling techniques to investigate whether an abstract phone could emerge from exposure to speech sounds. In effect, the study represents an attempt for operationalize a theoretical device of Usage-based Linguistics of emergence of an abstraction from language use. Our quest focuses on the simplest of such hypothesized abstractions. We test two opposing principles regarding the development of language knowledge in linguistically untrained language users: Memory-Based Learning (MBL) and Error-Correction Learning (ECL). A process of generalization underlies the abstractions linguists operate with, and we probed whether MBL and ECL could give rise to a type of language knowledge that resembles linguistic abstractions. Each model was presented with a significant amount of pre-processed speech produced by one speaker. We assessed the consistency or stability of what these simple models have learned and their ability to give rise to abstract categories. Both types of models fare differently with regard to these tests. We show that ECL models can learn abstractions and that at least part of the phone inventory and grouping into traditional types can be reliably identified from the input.

Error-Correction Learning; Memory-Based Learning; Widrow-Hoff rule; Temporal Difference rule; phone; abstraction; English

---

[*] Corresponding author: Petar Milin <p.milin@bham.ac.uk>

## 1. Introduction

Many theories of language presuppose the existence of abstractions that aim to organize the extremely rich and varied experiences language users have. These abstractions are thought to be either innate (generative theories) or to arise from experience (emergentist, usage-based approaches). Several mechanisms or principles by which abstractions could form have been identified. The mechanisms that are of particular interest for this study include: grouping similar experiences together, forgetting some dimensions of experience, and not attending to certain information streams to manage the influx of information. It is important to keep in mind that these mechanisms should not be considered as mutually exclusive, and neither is the list comprehensive: it focuses on principles that are attractive for linguistic theorizing and research (see, for example, Milin, Divjak, Dimitrijević, & Baayen, 2016).

In addition, we consider another mechanism to investigate whether learning can account for the emergence of linguistic abstractions from exposure to the ambient language: learning to ignore, unlearning or, more technically, filtering uninformative cues to form abstract knowledge. This principle or mechanism appears similar to forgetting, but while the former assumes an active process, the latter is more passive or spontaneous. Like the principle of grouping, there is a computational framework for filtering and both have shown considerable success in modeling a wide range of phenomena in linguistics and language cognition (for an application of the grouping principle, see TiMBL, Daelemans & Van den Bosch, 2005; for filtering, Baayen, Milin, Đurđević, Hendrix, & Marelli, 2011 presented their discriminative learning framework that is based on the Rescorla-Wagner 1972 rule, which is virtually identical to the Widrow-Hoff 1960 rule long applied in filtering problems).

We illustrate these mechanisms using a case study on the sounds of English in which we computationally model whether an abstract phone could emerge from exposure to speech sounds under supervised learning. This resembles the real-life task young children face when learning to read or write: a key part of this process lies in learning to recognize phonemes and mapping those onto graphemes. We simulate this process using three cognitively plausible learning rules: Memory-based learning (MBL), Widrow-Hoff (WH) and Temporal Difference learning (TD; a direct generalization of WH). The latter two computational models learn to filter irrelevant dimensions of experience via the cue competition inherent in the experience. They focus on those dimensions that help deal with the environment on a 'good enough' basis. What remains would then represent an abstraction: it is less detailed (or more schematic) than the input exemplars and more parsimonious because, as we learn, we discard or ignore what is non-predictive, i.e., not helpful for improving performance (by error-correction), and we retain only the relevant 'core'. Importantly, however, that abstraction is not a given and static memory trace, but rather an ever-evolving information-rich residue of the experience (cf., Love, Medin, & Gureckis, 2004; Nosofsky, 1986; Ramscar & Port, 2016).

*1.1 Much ado about abstractions*

If there is one thing that linguistic theories have in common, it is that they resort to abstractions for describing language. Some of these abstractions are so commonly agreed upon that they are taught to children in elementary school (think of parts of speech like noun, verb, or adjective). Others are specific to a particular theory (such as constructions).

Depending on the theory, these abstractions are either innate and present at birth, as is the case in generative frameworks, or they have emerged from exposure to the ambient language, as is assumed in usage-based approaches (see Ambridge & Lieven, 2011, for an overview of these theories and the predictions they make regarding language acquisition or development).

This implies that abstractions play a different role in both types of theories. In earlier versions of generative linguistic theories, abstractions were underlying representations: they are part of the deep structure which harbors the abstract form of a sound, word or sentence before any transformational rules have been applied. In usage-based approaches, abstractions are posited at the other end. They are built up gradually, as experience with language accumulates and commonalities between instances are detected. It is not uncommon for usage-based approaches to assume redundant encoding in which abstractions are stored alongside exemplars (Goldberg, 2006; Langacker, 1987). In fact, both approaches allow abstractions to co-exist with actual usage. The difference is in the sequence: on a generative approach, in a sense, abstractions give rise to usage while on a usage-based approach, usage gives rise to abstractions.

This practice of describing language using abstractions has led to the assumption that language knowledge and use must involve knowledge of the relevant abstractions (but see Ambridge, 2020). The challenge facing those who assume the existence of abstractions is one of proposing a hypothesized construct or mechanism that would give rise to these abstractions. For nativists this means explaining how these abstractions could be present at birth, and for emergentists this means explaining how such abstractions can emerge from exposure to usage. Ideally this needs to be done in sufficient detail for the account to be implemented computationally and/or tested empirically.

*1.2 The taming of the shrew (the smallest of language units)*

Researchers have long searched for the smallest meaningful unit in language (Sapir, 1921; Trubetzkoy, 1969). The main goal of the search for an abstraction was to discover a basic building block on which linguistic theories can build (much like the atom, electron and Higgs Boson, and subsequent research in particle physics: Gillies, 2018).

In the study of sound, over the years, many different levels of abstraction have been proposed such as the mora, demisyllable, phoneme, syllable, or phonetic features (Goldinger & Azuma, 2003; Marslen-Wilson & Warren, 1994; Savin & Bever, 1970). These are all assumed to have mental correlates. The most commonly proposed and assumed building block or unit of abstraction is the phoneme or syllable (Goldinger & Azuma, 2003). Generally, phonemes are defined as abstractions of actual speech sounds that ignore or 'abstract away' details of the speech signal that do not contribute to differentiating meaning. Syllables as a basic unit are usually considered a combinations of speech sounds that carry a rhythmic weight and are used to differentiate meaning.

In this study, we will focus on phoneme-like units. The phonemic perception and representation of speech is arguably one of the best-established constructs of linguistic theory (for a review see Donegan, 2015, pp. 39-40). Our choice may appear as fortunate and unfortunate, simultaneously and in almost equal measure: on the one hand, the definition of phonemes as abstracted from the speech signal makes them particularly convenient to test the filtering mechanism while, on the other hand, phonemes have come under fire more than once in recent history. Port (2010) claims: "Phones and phonemes, though they come very readily to our conscious awareness of speech, are not valid empirical phenomena suitable as the data basis for linguistics since they do not have physical definitions." Work by Goldinger and Azuma (2003) investigated an Adaptive Resonance Theory (Grossberg, 2013) approach to the emergence of perceptual 'units' or self-organizing dynamic states. It argues that the search for a fundamental unit is misguided and advocates an exemplar representation. A recurring claim in the argument against the cognitive reality of phonemes is the richness of memories of experiences, including experiences of sounds (compare with Port, 2010). Although we do not deny the validity of the findings reported (e.g., Goldinger & Azuma, 2003; Port, 2007; Port & Leary, 2005; Savin & Bever, 1970), we do not either accept them as evidence that there cannot possibly be any cognitive reality to phoneme-like units. Instead, we consider these findings a mere indication of the fact that phoneme-like units are not necessarily the whole story. Compare here the findings of Buchwald and Miozzo (2011), who found that two systems need to be recognized to account for errors in spoken language production, i.e., a processing stage with context-independent representations and one with context-dependent representations.

As mentioned earlier, the theoretical need for a basic building block in linguistics stems from a desire to taxonomize or systematize a complex system. Furthermore, this exercise is assumed to have (direct and indirect) benefits for a learner. Much of the early speech-error work provides strong evidence for the need for such representation (e.g., Fromkin, 1971). The lack of invariance problem from speech perception research illustrates this need nicely. Simply put, the acoustic signal is highly complex and variable and there is no simple mapping from the acoustic signal to some sort of phonetic structure (Appelbaum, 1996; Shankweiler & Fowler, 2015). For example, acoustic cues, like formant transitions differ depending on what the following vowel is (cf., Liberman, Cooper, Shankweiler, & Studdert-Kennedy, 1967; for production see Mielke, Baker, & Archangeli, 2016). Nevertheless, while the lack of invariance problem has advanced research in the field of speech perception and has not only increased our understanding of speech production and perception but has also been influential in early attempts at speech synthesis, not everyone agrees that it is as a big of a problem as previously thought (e.g., Appelbaum, 1996; Goldinger & Azuma, 2003). Even so, the search for a basic building block which linguistic theories can use as a concept on which to build additional linguistic structure continues.

It is also worth noting that the main motivation for our choice is not conceptual but methodological. This is to say that, although we are not opposed to the existence of phoneme-like units, neither do we believe that phonemes are the only abstract form cognitive representations of sound can take. Abstractions are likely language-dependent and could range from what might traditionally be referred to as moras, allophones, phonemes, demi-syllables, syllables, words and sequences of words. A productive theoretic proposition should not commit a priori to particular constructs or abstractions. In that respect, such units cannot be compulsory and the nature of these units could vary across languages or even individuals. Furthermore, such units could lack support in a cognitive sense and, yet, still be didactically useful (hypothetical) constructs. Finally, we do not consider these abstractions or units to be static either, as we address the question of how they unfold over time.

The main goal of this study, hence, is to explore whether abstract units can be emergent from input, using computational models that are well-known and established in the fields of linguistics and the psychology of language. These models are not only simple and therefore easily tractable but also biologically or cognitively plausible. For the remainder of the article we will distinguish between phonemes and phones and use phones to refer to the segments we are working with, to reiterate the fact that we are using phoneme-like units out of convenience and not out of any theoretical leaning.

*1.3 To learn or not to learn*

For the purposes of this study, we zoom in on a specific subset of linguistic theories and computational frameworks (i.e., models, algorithms) commonly used to support these theories. Linguistically, we will assume that universal sound inventories are not available at birth but are learned using general cognitive mechanisms. Psychologically, we will contrast models that assume veridical encoding of the input with those that allow or even encourage information sifting. Translated into the terms of the relevant machine learning approaches, this opposition would pitch grouping against filtering principles; more concretely, Memory-Based Learning (MBL) against Error-Correction Learning (ECL). Of course, much more sophisticated models are available and detailed neural mechanisms for learning and inference in complex probabilistic models, claimed to be biologically and cognitively plausible, have been proposed (e.g., Beck et al., 2008; Boerlin, Machens, & Denève, 2013; Buesing, Bill, Nessler, & Maass, 2011; Haefner, Berkes, & Fiser, 2016; Vértes & Sahani, 2019). Our selection of algorithms is based on the linguistically appealing and, thus, popular choice (MBL) and the simplicity of the competitor to that popular choice (ECL), which is gaining in popularity, particularly in research on word processing. ECL is additionally interesting since its simplest representative, the Widrow-Hoff or Delta rule, constitutes the cornerstone of many other, more powerful and more complex learning models (e.g., many connectionist models: McClelland & Rumelhart, 1989; also see discussion in Widrow & Lehr, 1990). Thus, our simple ECL can serve as 'proof of concept': with bells and whistles, performance should get better, but will also become gradually more intractable. Yet tractability is one of our main concerns: we want to be able to pinpoint how abstract linguistic knowledge would emerge, and under which conditions this process would falter.

*1.3.1 Capitalising on storage: MBL*

Exemplar models were not developed especially for language; rather, they apply equally to linguistic and non-linguistic phenomena. An exemplar model is therefore also not, in itself and on its own, a theory of language; rather, it is a model of memory representation (Divjak, 2019, p. 43) or, even more broadly, of how an entity (a human, an animal, a machine) accounts for new information. For language research, specifically, such models embody the usage-based view that language is part of general cognition and they allow us to access explanations for linguistic phenomena by venturing outside language, into the domain of general cognition.

Exemplar models propose that memory for linguistic experience is like memory for other experiences (Bybee, 2013, p. 52). Memory storage for linguistic experience is said to contain detailed information about the tokens that have been processed, including their form and the contexts in which they were used. Each instance of experience with language has an impact on its cognitive representation and when stored representations are accessed during encoding or decoding, the representations themselves change. Langacker (2017) rightly remarks that exemplars cannot be stored – they are transient neural events. What is 'stored' is a trace of their occurrence as each occurrence adjusts synaptic connections.

Linguistic exemplars come in a variety of sizes, ranging from a single segment, such as a vowel, to longer stretches of text, such as poems. Tokens of experience that are judged to be the same are grouped together and form exemplar clouds. These exemplar clouds constitute the categories of a language and linguistic ability resides in established patterns of activity with varying degrees of entrenchment. Exemplar categories are structured by similarity and frequency (Nosofsky, 1988) and often exhibit prototype effects (Rosch, 1975) because of differences in degrees of similarity. Note that an individual exemplar ‑ which is a detailed perceptual memory ‑ does not correspond to a single perceptual experience, but rather to an equivalence class of perceptual experiences (Pierrehumbert, 2001; but see, for example Ferro, Marzi, & Pirrelli, 2011, for a 'memory greedy' self-organizing model of word storage and processing). Exemplars can thus differ in strength depending on the number of tokens they contain: exemplars built up from a large number of tokens will be represented more strongly than those built up from a smaller number of tokens. The stronger exemplar or set of exemplars often forms the centre of a category and other exemplars are more or less similar to that stronger exemplar or set of exemplars. The exemplar cloud of a word would include all the meanings and contexts in which the word has been experienced. This 'excessive' storage is believed to be possible thanks to our impressive neural capacity and because information is stored in a highly structured and efficient way, i.e., by forming categories (even if only as and when needed, see Nosofsky, 1988) and storing similar items in proximity to one another (Ferro et al., 2011).

On an exemplar model of language, cognitive representations are built up as language users encode utterances and categorize them on the basis of sound, meaning and context. As incoming utterances are sorted and matched by similarity to existing representations, units such as syllable, word and construction are thought to emerge. Linguists are particularly interested in how much generalization and abstraction occurs in the organization of linguistic experience. Exemplar representations contrast at all levels – phonetic, morpho-syntactic, semantic and pragmatic – with the more abstract representations of structural or generative theories, because variation and properties predictable from general principles are retained in exemplar models. Exemplar accounts are particularly well established in phonetics and phonology (e.g., Johnson, 1997; Goldinger, 1998; Bod, 1998; Pierrehumbert, 2001; Hay & Bresnan, 2006), yielding a situation in which words or phrases are stored with rich phonetic detail. However, this opinion, although widespread, does not seem to mesh with the empirical findings from studies on echoic sensory memory, representing the short-term storage of auditory input (cf., Jääskeläinen, Hautamäki, Näätänen, & Ilmoniemi, 1999): sensory memory has been found to be very short-lived, and hence it remains unclear how that rich detail would make it into long-term memory (under normal, non-traumatic circumstances).

Technically, most (if not all) MBL approaches store past experiences as correct input-output pairs: the input is a (vector) representation of the environment and the system, whereas the output assumes an action or an entity – a desired outcome (see Haykin, 1999). Memory is organized in neighborhoods of similar or proximal inputs, which presumably elicit similar actions/entities. Thus, a memory-based algorithm consists of (a) a criterion for defining the neighborhoods (similarity, proximity) and (b) a learning principle that guides assignment of a new exemplar to the right neighborhood. One simple yet efficient example of such a principle is that of Nearest Neighbors, where a particular type of distance (say, the Euclidean distance) is minimal in a given group of (near) neighbors. It has been shown that this simple principle straightforwardly ensures the smallest probability of error over all possible decisions for a new exemplar (as discussed in Cover & Hart, 1967). The principle appears simple but it is also efficient. Moreover, it appeals to linguists, and Cognitive Linguists in particular, as it equates learning to categorization or classification of newly encountered items into neighborhoods of 'alikes': new items (exemplars) are processed on the basis of their similarity to the existing neighborhoods, which resembles the process of analogical extension. And for Cognitive linguists, categorization is one of the foundational, general cognitive abilities that facilitate learning from input (Taylor, 1995).

*1.3.2 Making the most of filtering: ECL*

The mechanisms of ECL have been enthusiastically received through early work on biologically inspired Artificial Neural Networks, in the work of Hebb (1949), Rosenblatt (1958), and Widrow and Hoff (1960), in particular. Although they have had great success in various practical applications, their further development was stalled and finally halted, partly due to success in the development of serial computers and partly because of the harsh criticism these models received from Minsky and Papert (1969).

Almost completely independently, Rescorla and Wagner (1972) were working on a formal (i.e., mathematical) model of Classical or Pavlovian conditioning in animals and humans. Their research built on seminal ideas in the work of Hull (1943) and the linear model of Bush and Mosteller (1955). The critical change that Rescorla and Wagner introduced was that of association with all relevant learning cues and not just a particular one, as proposed by Hull. The idea of such a *total associative strength* made the Rescorla-Wagner learning rule effectively identical to the rule of Widrow and Hoff (cf., Anderson, 2000; Rescorla, 2008). Hence, by extension, the Rescorla-Wagner rule itself is based on the process of error-correction and the principle of filtering (cf., Haykin, 1999). From a basic psychological point of view, the contemporary take on the Rescorla-Wagner rule, specifically, and ECL, more generally, focuses on (a) the total associative strength and the implication of it that learning is, almost always based on (b) a compound of cues, emerging through (c) cue competition and with (d) the general role of context (cf., Anderson, 2000; Bouton, 2007).

The Rescorla-Wagner rule became one of the most successful and most appraised models of animal and human learning: a benchmark rule to compare other models with (for an excellent recent example, see Kokkola, Mondragón, & Alonso, 2019). In language research, the Rescorla-Wagner model was introduced in the work of Ellis (2006a) on second language learning, and Ramscar et al. (2007; 2010) on language acquisition. Finally, the Naïve Discrimination Leaning (NDL: Baayen et al., 2011) computational framework made it possible to 'think big' as it allowed for scaling up and modelling large amounts of textual data across the broad range of languages and language processing tasks.

Bouton (2007) pointed out how 'entertaining' the Rescorla-Wagner can be as it leads to a variety of peculiar predictions. This is exactly what seemed to have inspired Ellis (2006a) to ask the question of whether L2 acquisition is 'irrational', and to convincingly argue that many of the apparently odd phenomena that can be observed in L2 effectively fall out from the principles of associative learning theory. Ellis, first, lays out the general dynamics which are based on these general principles of learning: typically, human learning takes place in complex situations that are important as learning setting or learning background (viz. a compound of cues; also see Divjak, Milin, Ez-zizi, Józefowski, & Adam, 2021; Romain, Ez-zizi, Milin, & Divjak, 2022) where a multitude of cues compete for predictive association with (or discrimination of) an outcome. New information comes in against the backdrop of our accrued experience with these cues, outcomes, and their setting. These complex dynamics lead Ellis to assert further that many oddities or 'irrationalities' of L2 learning occur exactly because L2 learning does not begin from a blank slate, but typically when considerable L1 experience has already accrued; this then shapes the later L2 intake (cf., Ellis, 2006b; also, see Ellis, 2016, for further discussion of these points from the point of view of Complex Adaptive Systems). This theorizing makes ECL 'organically' appealing to Usage-based linguistics, with its cognitive commitment and interest in emergentism (cf., Divjak, 2019).

The work of Ramscar and his collaborators, conversely, focused on L1 acquisition from a, what they call, discriminative point of view. The term "discriminative" is used to describe the mechanism that is implemented in error-driven learning models: the error-driven mechanism triggers cue competition which is said to "re-represent an input representation so as to maximize its informativity about a set of outputs" (see, in particular, Ramscar & Yarlett, 2007; Ramscar et al., 2010; Arnon & Ramscar, 2012). This particular interpretation of error-driven learning is reminiscent of James (1890) view, in that it assumes that, at birth, the mind consists of a set of undifferentiated input and output states, which are gradually differentiated. For Ramscar and his collaborators, error plays an important role in this process of individuation.

NDL is a computational modelling framework for studying language at scale (Baayen et al., 2011; Milin et al., 2016). Based on the Rescorla-Wagner rule (1972), it implements a simple two-layer network of input and output units which are all connected. From a linguistic perspective, its particular attractiveness for language research is not just because it allows engaging with larger corpora, but also for its tractability and wide range of language phenomena that can be successfully modelled. These phenomena include frequency and family size effects, contextual and paradigmatic effects on the processing of simple, inflected and compound words, and many others, across a range of experimental tasks and different languages. In recent years, the NDL framework has branched out. On the one hand, computationally, NDL evolved into the more potent Linear Discrimination Learning (LDL: Baayen, Chuang, Shafaei-Bajestan, & Blevins, 2019), which operationalizes constructs from Word-and-Paradigm morphology (e.g., Blevins, 2016) and can handle auditory processing tasks (e.g., Shafaei-Bajestan, Moradipour-Tari, Uhrig, & Baayen, 2021). On the other hand, NDL extended to and integrated with Usage-based linguistics, and Cognitive linguistics in particular. Examples include the use of Behavioural Profiles as learning cues (Milin, Divjak, & Baayen, 2017), and a closer investigation of what linguistic knowledge looks like if we allow it to emerge from exposure to usage (Divjak et al., 2021; Romain et al., 2022).

ECL runs on very different principles than MBL. It filters the value (weight) of the input information (signal) to facilitate error-free prediction of some desired outcome. The process unfolds in a step-wise manner as new input information becomes available. The weighting or evaluation of the current success is done relative to a mismatch between the current prediction and the true or desired outcome. In other words, the stimulation to modulate the operating parameters comes from the difference between the desired outcome and the current prediction of it on the basis of available input data. That difference induces a rather small disturbance in the input weights which, effectively, evaluates the input data itself to bring its prediction closer to the true desired outcome. The learning process continues as new data arrives. Gradually, the difference between the prediction and the true outcome becomes smaller and, consequently, the changes in the input weight also diminish over time. Fitz and Chang (2019) have identified ERP evidence for prediction error in language learning: their connectionist model can simulate data from studies on the P600 (agreement, tense, word category, subcategorization, and garden-path sentences), among other phenomena. Anderson, Holmes, Dell, and Middleton (2019) established experimentally that learning in the phonological domain is greater when productions are more errorful, thus demonstrating that learning of phonotactic patterns specifically is triggered by the need for error correction.

### 1.3.3 Contrasting MBL and ECL

We can safely assume that the smallest common denominator of the two approaches is the ability to learn from the environment for the sake of ever better adaptation. Under such assumption, rather simplified, a system (be it a computer, animal or human) learns about its environment by processing input information about the environment and about itself in relation to that environment (viz. Rescorla, 1988). Gradually, that system becomes more knowledgeable about its environment and, consequently, better adapted to it. More precisely, however, MBL and ECL could be seen as opposing principles: in the extreme, an exemplar approach does not require anything beyond organized (i.e., grouped) storage, while an error-correction approach is a stockpile of perpetually changing associations.

To researchers interested in linguistics, and cognitive linguistics in particular, Memory-Based Learning (MBL) appeals because it explains linguistic knowledge in terms of categorization and analogical extension. However, Error-Correction Learning (ECL) appears to be conceptually better suited for testing the emergentist perspective on language knowledge as it is directly revealing about the process itself – the emergence of knowledge (see Julià, 1983 about the importance of understanding the process and not just the end-state for linguistic theorizing), and can take a holistic (i.e., molar) approach to knowledge (e.g., a long tradition, going back to Tolman, 1932; Osgood et al., 1954; and more recently to the ideas of Rescorla, 1988; Ellis, 2006a; Ramscar et al., 2010 etc.). This knowledge evolves gradually and continuously from the input which is conceived of as a multidimensional space of 'discernibles' or contrasts. In other words, rather than retaining the exemplars as they are, in all their unparsimonious richness, learners may disregard non-predictive aspects of the input, keeping only what is *useful* in an adaptive sense.

It follows that, computationally, learning assumes stimulation from the environment to which the system reacts by modulating its parameters (also known as free parameters) for operating in and on that environment (cf., Mendel & McLaren, 1970). Ultimately, the system responds to its environment in a new and more adapted way. From this computational perspective, what makes these two learning processes unique is the specific way in which parameter change happens. Generally speaking, MBL stores or memorizes data directly into neighborhoods by similarity, while ECL filters that data to retain what is necessary and sufficient.

Coincidentally, both MBL and ECL are currently being pushed to their (almost paradoxical) extremes. On the one hand, Ambridge (2020, https://osf.io/5acgy/) denounces any stored linguistic abstractions (while acknowledging other types of abstractions). On the other hand, Ramscar (2019) denies language units any ontological (and, consequently, epistemological) relevance by disputing mappings of any substance (and, hence, theoretical importance) between form units and semantic units: units become 'dis-unit-ed' due to a lack of one-to-one mappings between an (allegedly fixed) form and an (allegedly fixed) meaning.

*1.4 This study: pursuing that which is learnable, with some caveats*

In this paper, we contrast a model that stores the auditory signal in memory with one that learns to filter (i.e., discriminate) phones in the auditory signal. Modeling whether an abstract phone could emerge from exposure to speech sounds by applying the principle of error-correction learning crucially relies on the concept of unlearning, whereby the strength of the links between any cues that do not co-occur with a certain outcome are reduced (see Ramscar, Dye, & McCauley, 2013, for discussion). The principle is such that the model will gradually try to unlearn irrelevant variations in the speech input, discard or filter out dimensions of the original experience and retaining more abstract and parsimonious 'memories'. This strategy is the radical opposite of a model that presupposes that we store our experiences veridically, i.e., in their full richness and idiosyncrasy. On such an account, learning equals encoding, storing and organizing that storage to facilitate retrieval.

To achieve this, we will provide our models with a common input in automatic speech recognition, Mel Frequency Cepstral Coefficients (henceforth MFCCs), derived from the acoustic signal, and with phones as outcomes. We will assign our models the task of storing vs. learning to associate the two. This implies that, in a very real sense, our modeling is 'artificial': our model will learn phoneme-like units because those are the outputs that we provide. If we provided other types of units, other types of outputs might be learned. Yet, this does not diminish our finding that phones can be successfully learned from input, using an error-correction learning approach.

Of course, because we use a simple algorithm, performance differences with state-of-the-art classifiers should be expected (e.g., Graves & Schmidhuber, 2005). Yet, what we trade in, in terms of performance, we hope to gain in terms of interpretability of the findings and continuation with productive language research traditions that make use of MBL and/or ECL. Rather than maximizing performance, we will minimize two critical 'investments': in cultivating the input, and in algorithmic sophistication (and computational intractability). In other words, we remain as naïve as possible and engage with MFCCs directly as input, and we use one of the simplest yet surprisingly efficient computational models of learning. Finally, recall also that we do not claim that phonemic variation would be something, let alone everything, that listeners identify. In that sense, our methodology consists of a carefully chosen set of worst-case scenarios, i.e., computationally challenging learning situations that would assume hearing one speaker only, or having limited exposure to different words, or learning from data with strictly random noise. We will analyze the differences in performance of selected learning algorithms under such worst-case scenarios. This should add to the body of knowledge in, at least, two important ways: first, we will establish the learnability of phones by the simplest, theoretically well-understood learning algorithms; second, we will present the details of the performance of said algorithms. In other words, we will advance our understanding of both the data and the learning mechanisms. Recall that the learning models or algorithms are chosen either because they are rooted in the experimental tradition of biology and psychology and are well-suited to support emergentism (ECL), or because they represent an attractive choice for linguists as they build on frequency, memory, and analogy (MBL). Yet, they are certainly not competitors to the state-of-the-art computational systems.

As said, learnability is our minimal requirement *and* our operationalization of cognitive reality. On a usage-based, emergentist approach to language knowledge, abstractions must be learnable from the input listeners are exposed to if these abstractions are to lay claims to cognitive reality (Divjak, 2015a; Milin et al., 2016). Whereas earlier work explored whether abstract labels can be learned and whether they map onto distributional patterns in the input (cf., Divjak, Szymor, & Socha-Michalik, 2015), more recent work has brought these strands together by modeling directly how selected abstract labels would be learned from input using a cognitively realistic learning algorithm (cf., Milin, Divjak, et al., 2017).

As we will explain in Section 2, we ran our simulations on much simpler learning algorithms. In other words, Schatz, Feldman, Goldwater, Cao, and Dupoux (2021) concluded that plausible or realistic data can generate a learnable signal (i.e., in terms of the joint probability distribution, given a problem space), and we now move to make the crucial step to simultaneously test (a) plausible learning mechanisms against (b) realistic data. Sceptics might argue that an explicitly supervised learning scenario for the acquisition of phonetic abstraction is, in itself, cognitively unrealistic. We would agree with this claim, in principle, at least if the irrelevance is established from the point of view of cognitive (neuro-) science and language learning in the wild. But that does not make the task irrelevant to our lives (see, for example, Graves & Schmidhuber, 2005). The process of learning to read includes, for example, learning to identify phonemes and to syllabify the printed words (e.g., Nesdale, Herriman, & Tunmer, 1984; Tunmer & Rohl, 1991).

*1.5 Cognate work*

An error-correction approach to speech sound acquisition has been attempted. Nixon and Tomaschek (2020) trained a Naïve Discriminative Learning model (NDL), relying on the Rescorla-Wagner learning rule, to form expectations about a subset of phones from the surrounding speech signal. For this study, data from a German spontaneous speech corpus was used (Arnold & Tomaschek, 2016). The continuous acoustic features were discretized into 104 spectral components per 25 ms time-window, which were then used as both inputs and outputs. A learning event consisted of the two preceding cues (104 elements each), a target (104 elements) and one following cue (also 104 elements). Learning proceeded in an iterative fashion over a moving time-window. The trained model was tested by trying to replicate an identification task on vowel and fricative continua, and was able to generate reassuring discrimination curves for both continua. In other words, Nixon and Tomaschek (2020) investigated sounds that are discriminated based on cues from the expected spectral frequency ranges for vowels and fricatives.

While the present study is similar to the work by Nixon and Tomaschek (2020), both in spirit and in substance, it also differs in several major ways. First, instead of predicting learned activations, we used activations as predictors of the correct phone. Second, the present study uses the entire phone inventory of English and does not focus on a subset. Third, we use real-value input from MFCCs directly, rather than transformed into discrete categories. That is why we rely on the Widrow-Hoff learning rule, noting that it is considered to be identical to the Rescorla-Wagner rule (as Rescorla, 2008, himself, pointed out) but accepts numeric input (see Milin, Madabushi, Croucher, & Divjak, 2020, for further details). Fourth, our approach, specifically, is grounded in previous empirical work within the NDL framework (cf., Milin, Feldman, Ramscar, Hendrix, & Baayen, 2017), and uses two learning measures: directly proportional activation from the input MFCCs, and inversely proportional diversity of the same input MFCCs. The former indicates the support that an outcome (in this particular case a phone) receives from the input, while the latter expresses the competition among outcomes activated by the same input cues (for details, see, Milin, Feldman, et al., 2017). Finally, Nixon and Tomaschek (2020) focus on the predictivity of the resulting model, i.e., on predicting responses in discrimination and categorization tasks, and less on the creation of abstract units. The authors do, however, discuss interesting initial findings that begin to replicate categorical perception results from speech perception.

It is important to point out that our study does not contradict recent developments within the discrimination-learning framework to language. The Linear Discriminative Learner (LDL), in particular, sets itself the more ambitious goal to refute the need for hierarchies of linguistic constructs (cf., Baayen et al., 2019; Chuang & Baayen, 2021; Baayen, Chuang, Shafaei-Bajestan, & Blevins, 2019; Shafaei-Bajestan et al., 2021).

The present study, however, asks a more fundamental question relating to the learnability of the simplest of such linguistic abstractions – phones. Yet, in that, we take a radical approach and model the emergence straightforwardly from continuous, numeric input representations (i.e., MFCCs), rather than from pre-processed discrete units (i.e., FBSF of Arnold, Tomaschek, Sering, Lopez, & Baayen, 2017, as used in the related studies by Nixon & Tomaschek, 2020; Shafaei-Bajestan et al., 2021; also see Nenadić, 2020).

Our learning-based approach is also reminiscent of the mechanism-driven approach proposed in Schatz et al. (2021). Their findings suggest that distributional learning might be the right approach to explain how infants become attuned to the sounds of their native language. The proposed learning algorithm did not try to achieve any biological or cognitive realism but focuses purely on the power of generative machine learning techniques (see Jebara, 2012, for details on this family of Machine Learning models, and the distinction between discriminative and generative learning model families). Thus, they selected an engineering favorite which is based on a Gaussian Mixture Model (GMM) classifier. The continuous sound signal was sampled in regular time-slices to serve as input data to generate a joint probability density function which is a mixture of (potentially many) Gaussian distributions from the given samples. The focus of the study by Schatz et al. (2021) was, to reiterate, not on the plausibility of a particular (generative) learning mechanism, but rather on the realism of the input data and its potential to be learned. This justifies their choice for a powerful generative learning model.

## 2. Methods

One of the challenges of working with speech is that it is continuous. In this section we describe the speech materials and how they were pre-processed before we move on to presenting the details of the modeling procedures. Recall that our decision to model phones does not reflect any theoretical bias on our part: we believe that similar results would be obtained with any number of other possible levels of representation.

*2.1 Speech Material*

The speech items used for modeling stem from the stimuli in the Massive Auditory Lexical Decision (MALD) database (Tucker et al., 2019). This database contains 26,000 word stimuli and 9,500 pseudo-words produced by a single male speaker from Western Canada; together, this amounts to approximately 5 hours of speech. The forced alignment was completed using the Penn Forced Aligner (Yuan & Liberman, 2008). The forced-aligned files provide time aligned phone-level segmentation of the speech signal often used in phonetic research and for training of automatic speech recognition tools. All phone-level segmentation is indicated using the ARPAbet transcription system (Shoup, 1980) for English, which we use throughout this paper. The speech stimuli, along with force-aligned TextGrids, are available for use in research (http://mald.artsrn.ualberta.ca).

In order to prepare the files for modeling we first set the sampling frequency to 16 kHz. We then windowed and extracted MFCCs for all of the words using 25 ms long overlapping windows in 10 ms steps over the entire acoustic signal. By way of example, if there was a 100 ms phone it would be divided into 9 overlapping windows with some padding at the end to allow the windows to fit. MFCCs are summary features of the acoustic signal used commonly in automatic speech recognition. MFCCs are created by applying a Mel transformation, which is a non-linear transform meant to approximate human hearing, to the frequency components of the spectral properties of the signal. The mel frequencies are then submitted to a Fourier transform, so that, in essence, we are taking the spectrum of a spectrum known as the cepstrum. Following standard practice in automatic speech recognition, we used the first 13 coefficients in the cepstrum, calculated the first derivative (delta) for the next 13 coefficients, and the second derivative (delta-delta) for the final set of coefficients (Holmes, 2001). This process results in each window or trial having 39 features, thus for each word we created an MFCC matrix of the 39 coefficients and the corresponding phone label. The selection of MFCCs was a convenience decision and future exploration of other types of input should be pursued. We note, however, that the focus of the present article is on properties of the learning algorithms, and on what is learned given those properties, rather than on the properties of the input.

*2.2 Algorithms*

As stated in the Introduction, MBL and ECL can be positioned as representing opposing principles in learning and adaptation, with the former relying on storage of otherwise unaltered input-outcome experiences, and the latter using filtering of the input signal to generate a minimally erroneous prediction of the desired or true outcome. These conceptual differences are typically implemented technically in the following way: the MBL algorithm must formalize the organization of stored exemplars in appropriate neighborhoods and then, accordingly, the assignment of new inputs to the right neighborhood. Conversely, the ECL algorithm would typically use iterative or gradual weighting of the input to minimize the mismatch between predicted and true (correct) outcome.

We applied MBL by 'populating' neighborhoods with exemplars that are corralled by the smallest Euclidean distance, and then assigning new input to the neighborhood of $k = 7$ least distant (i.e., nearest) neighbors. This is, arguably, a rather typical MBL implementation, as Euclidean distance is among the most commonly used distance metrics, while k-Nearest Neighbors (kNN) is a straightforward yet efficient principle for storing new exemplars (as discussed previously by Cover & Hart, 1967; Haykin, 1999 etc.). The neighborhood size was set to 7, following the findings from a study in which the neighborhood size was systematically varied (Milin, Keuleers, & Đurđević, 2011) and where $k = 7$ was suggested for optimal performance: small (and exclusive) enough to ensure good differentiation, but large (and inclusive) enough to provide a solid basis for assigning new exemplars.

Two ECL learning models, Widrow-Hoff (WH; Widrow & Hoff, 1960) and Temporal Difference (TD; Sutton & Barto, 1987, 1990), were chosen as two versions of a closely related idea: WH (also known as the Delta rule or the Least Mean Square rule; see Milin et al., 2020, and further references therein) represents the simplest error-correction rule; TD generalizes to a rule that accounts for future rewards, too. More precisely, at time step $t$, learning assumes small changes to the input weights $w_t$ (small to obey the Principle of Minimal Disturbance; cf., Milin et al., 2020) to improve prediction in the next time step $t + 1$. This is achieved by re-weighting the *prediction error* (which is the difference between the targeted outcome and its current prediction) with the current input cues and a *learning rate* (free parameter): $\lambda(o_t - w_t c_t)c_t$, where $\lambda$ represents the learning rate, $c_t$ is the vector of input cues, $w_t c_t$ is the prediction and $o_t$ is the true outcome. The full WH update of weights $w$ in the next time step $t + 1$ is then:

$$w_{t+1} = w_t + \lambda(o_t - w_t c_t)c_t$$

For TD, however, the prediction ($w_t c_t$) is additionally corrected by a future prediction as the (temporal) difference to that current prediction: $w_t c_t - \gamma w_t c_{t+1}$, where $\gamma$ is a *discount factor*, the additional free parameter that weights the importance of the future, where $w_t c_{t+1}$ is the prediction given the future – i.e., the next input $c_{t+1}$. Note that if the discount factor is set to $\gamma = 0$, which means that the future is completely irrelevant or unimportant, and the algorithm becomes identical with the Widrow-Hoff rule. The update rule for the TD thus becomes:

$$w_{t+1} = w_t + \lambda(o_t - [w_t c_t - \gamma w_t c_{t+1}])c_t$$

There are three further points that guided our choice of the three algorithms, and which require some elaboration. First, as we already pointed out, both MBL and ECL have had considerable success in modelling various language phenomena. They are well-known and popular among linguists and psychologists. Thus, the current results can be straightforwardly linked to a considerable body of existing work.

Second, we decided to apply the most commonly used/ default set-up and settings for any free parameters, and did not tune them to achieve better results. On the one hand, tuning will not help to achieve results that would compare with the state-of-the-art machine learning techniques (e.g., so-called deep learning models). On the other hand, changing free parameters would make the current findings less tractable, perhaps more idiosyncratic, and ultimately less directly relatable with findings from previous studies.

Finally, our decision to use two closely related ECL algorithms, WH and TD, is to allow for a detailed examination of the effect that our temporally overlapping data (see Section 2.1) might have on error-correction itself. If the data in the next (temporally overlapping) timestep is informative for prediction, then the TD rule, by explicitly making use of that next data (i.e., $\boldsymbol{w}_t\boldsymbol{c}_t - \gamma\boldsymbol{w}_t\boldsymbol{c}_{t+1}$), ought to show an advantage over *greedy* WH that does not consider any future data at all (in effect, if we set the discount factor $\gamma$ to zero, the whole correction for future predictions $\gamma\boldsymbol{w}_t\boldsymbol{c}_{t+1}$ becomes zero too, and TD becomes identical to WH). Even more explicitly, in what follows, we will compare the results of the two chosen ECL algorithms with both the original (or raw) data and simulated data with Gaussian properties, where the noise is purely random and, hence, future data is uninformative. This way we will engage with both the data (overlapping and random) and the ECL algorithms (WH and TD) that should reveal differences in what gets learned from such data (see, for example, Sutton & Barto, 1990 for original work on TD as a generalization of the Rescorla-Wagner rule and, thus, of the Widrow-Hoff rule; for a more recent overview see Gershman, 2015).

*2.3 Training*

We ran several computational simulations of learning the phones of English directly from the MFCC input. This constitutes a 40-way classification task, predicting 39 English phones and a category of silence.

The full learning sample consisted of 2,073,999 trials that were split into training and test trials (90% and 10%, respectively). Each trial represents one 25ms window extracted from the speech signal and matched with the correct phone label. The first training regime used the training data trials as they are: at each trial, input cues were represented as an MFCC row-vector, and the outcome was the corresponding 1-hot encoded (i.e., present) phone. The time steps overlapped. The second training regime used the same 90% of training data to create $N = 100$ normally distributed data points (Gaussian), with estimated mean and 99% ($+/-2.58SE$) confidence intervals (to the true, population mean), one for each of the 40 phone categories, given the input MFCCs. The Raw input consisted of 1,866,599 trials (90% of the full dataset), while the Gaussian data contained 4,000 trials (100 normally distributed data points for each of the 39 phones and silence). Recall that, on the one hand, the two dataset types were introduced to see how much actual gain in learning we observe when the data is overlapping or temporally informative. On the other hand, this test also allows for understanding the nature of noise in the data better by testing whether it is safe to assume that it is (nearly) Gaussian. In other words, this comparative modelling exercise will reveal important information about both the learning algorithm and the data. As for the former, if the temporal order and overlap in Raw input is important and valuable, then this ought to be reflected as an advantage for TD vs. WH. As for the latter, we will learn more about how much of a difference such temporal information can make.

The two datasets – Raw vs. Gaussian input – were treated identically during training. For the two ECL algorithms the learning rate was set to $\lambda = 0.0001$, to safeguard from overflow, given the considerable range of MFCC values ($min = -90.41$, $max = 68.73$, $range = 159.14$). This learning rate was kept constant throughout all simulation runs. Additionally, when training with the Temporal Difference model, the decay parameter of future reward expectation was set to the commonly used value of $\gamma = 0.5$. Upon completion, each training yielded a weight matrix, with the MFCCs as input cues in rows and the phone outcomes in columns (39 × 40 matrix).

The training resulted in six weight matrices, given the two types of inputs (Gaussian and Raw) and the three learning models (Memory-Based, MBL, Widrow-Hoff, WH and Temporal Difference, TD). These matrices were used to predict phones in the test dataset ($N = 207,399$, or 10% of all trials). Raw MFCCs were fed to the algorithms as input and were weighted by the weights from the six respective matrices, learned under the distinctive learning regimes. We used weighted input sums, also called total or net input, as the current activation for each of the 40 phone outcomes, and the absolute length (1-norm) of the weighted sum indicated the diversity or the competition among the possible outcome phones (for details and prior use, see Milin, Feldman, et al., 2017). The prediction of the system was the phone with the highest ratio of activation over diversity, or support over the competition. This proposed simple ratio is based on previous research findings all of which have consistently shown a direct effect of activation and an inverse effect of diversity that is consequently explained as the measure of competition of relevant/possible outcomes given input cues (cf., Baayen, Milin, & Ramscar, 2016; Milin, Feldman et al., 2017; Divjak, Milin et al., 2021).

In addition to the ECLs we also trained a MBL with k-Nearest Neighbors, using Euclidean Distances and $k = 7$ neighbors. The memory resource was built from corresponding input-outcome data pairs: $N_{Raw} = 1,866,599$ pairs for Raw and $N_{Gaussian} = 4,000$ pairs for Gaussian data. Next, MFCC input cues were used to find the 7 nearest neighbors in the test data, and the outcome of a majority vote was considered as the predicted phone. That prediction was then weighted by the system's confidence (i.e., the probability of said majority vote against all other votes). For all MBL simulations we made use of the **class** package (Venables & Ripley, 2002) in the **R** software environment (R Core Team, 2020).

## 3 Results and discussion

The following four subsections present independent but related pieces of evidence for assessing the learnability of abstract categories, given MFCC input, and performance of the three chosen algorithms (MBL, WH, and TD) under different worst-case learning scenarios. Whether or not phones are learnable can be inferred from success in predicting new and unseen data from the same speaker, from both Raw (collected) MFCCs and Gaussian (generated), where the difference is in the presence/absence of temporal order and overlap (Section 3.1). Next, an arguably even more challenging task would be to predict unseen data from a new speaker, after learning from single-speaker data alone (Section 3.2). At the more general level, questions such as reliability and generalizability are of crucial importance. We will, hence, explore how consistent or reliable the success in predicting phones is across multiple random samples of the data (Section 3.3). Finally, it is important to understand whether what is learned generalizes vertically: do we see and, if so, to what extent do we see an emergence of plausible phone clusters, resembling traditional classes, directly from the learned weights (Section 3.4).

### 3.1 More of the same: predicting new input from the same speaker

We can consider our simulations as, essentially, a 3 × 2 experimental design, with as factors Learning Model (Memory-Based, Widrow-Hoff, and Temporal Difference) and Type of Input data (Raw vs. Gaussian). The success rates our three algorithms achieve in predicting unseen data, given Raw or Gaussian input, are summarized in Table 1.

Overall, the results reveal an above chance performance given both the naive or 'flat' probability of any given phone occurring by chance, which stands at 2.5% assuming an equiprobable distribution of the 40 outcome instances, or the probability of the phone occurring in the sample, which is, essentially, its relative frequency in the training sample. The error-correction learning rules (WH and TD) performed below the sampling probability only in a few cases, and only when using raw training data. Out of 40 instances, Widrow-Hoff underpredicted 12 phones, while Temporal Difference underpredicted 7 phones. MBL, on the other hand, always performed above the baselines set by chance and sampling probability.

The strength of the correlation between the probability of the phone in the test sample, on the one hand, and the success rates of the three learning models, on the other hand, adds an interesting layer of information for assessing the overall efficiency of the chosen learning models. To bring this out, we made use of Kendall's Tau-b ($\tau_B$), a Bayesian non-parametric alternative to Pearson's product-moment correlation coefficients. Table 2 contains these correlation coefficients and reveals two interesting insights. Firstly, despite the fact that, for all three learning algorithms, the Gaussian dataset did not reflect the sampling probabilities of the respective phonemes (100 generated data points per phone; N = 4000 total training sample size), the correlation with the phones' probabilities in the test sample increased considerably (importantly, the correlation between probabilities in the training and the test sample was near-perfect: $r = 0.99998$). In other words, even though the number of exemplars, i.e., the extent of exposure, was kept constant across all phones, the algorithms' successes resembled the likelihood of encountering the phones in the Raw samples, both for training and testing. Generally speaking, this then represents indirect evidence of learning rather than of simply passing through the equiprobable expectations (given that each outcome had 2.5% chance to occur).

Secondly, these correlation coefficients are considerably higher for Gaussian data than for their Raw counterparts. Likewise, the success rates for the two ECL algorithms increases. However, the opposite holds for the MBL algorithm: MBL shows a higher correlation for Gaussian data but combines this with a lower success rate when that same Gaussian data is used as stored exemplars.

Taken together, these two observations indicate that the ECL algorithms can do more with less. In addition, they are sensitive to the order of exposure or the trial-order effect in the case of Raw data: the Raw data was fed to the learning machines in a particular order of time-overlapping MFCC-phone pairs, while the Gaussian data was computer-generated and randomized. The incremental nature of ECL makes many models from this family deliberately sensitive to the order of presentation in learning. For example, the Rescorla-Wagner (1972) rule, the discrete data counterpart of the Widrow and Hoff (1960) rule, is purportedly susceptible to order. This is, in fact, the core reason why it can simulate the famous blocking-effect (cf., Kamin, 1969). Conversely, MBL simply stores all data in an as is fashion and is, consequently, blissfully unaware of any particular order in which the data comes in and, for that matter, does not assign any particular value to the order (assuming unlimited storage).

**Table 1.** Success rates of predicting the correct phone of three different learning models (Widrow-Hoff, Temporal Difference, and Memory-Based Learning) using two different types of input data (Raw or Gaussian). The leftmost column represents the probability of a given phone in the Raw training sample.

| Phone | Sampling Probability | Raw | | | Gaussian | | |
|---|---|---|---|---|---|---|---|
| | | MBL | WH | TD | MBL | WH | TD |
| silence | 8.03 | 54.26 | 41.58 | 52.33 | 58.19 | 34.58 | 34.51 |
| aa | 2.12 | 42.51 | 0.63 | 4.20 | 38.86 | 17.32 | 16.55 |
| ae | 2.64 | 52.82 | 19.78 | 30.02 | 51.66 | 29.02 | 31.26 |
| ah | 5.49 | 26.31 | 11.94 | 11.65 | 24.12 | 32.21 | 30.03 |
| ao | 1.02 | 38.24 | 5.38 | 8.33 | 23.13 | 19.15 | 19.34 |
| aw | 0.79 | 27.88 | 0.64 | 2.09 | 13.18 | 11.62 | 15.38 |
| ay | 2.58 | 41.03 | 0.00 | 10.14 | 28.35 | 23.83 | 26.13 |
| b | 1.22 | 46.46 | 11.24 | 7.50 | 16.96 | 23.42 | 24.92 |
| ch | 0.86 | 24.95 | 11.11 | 0.00 | 13.61 | 8.68 | 8.44 |
| d | 2.43 | 41.85 | 3.24 | 2.57 | 17.55 | 15.63 | 15.12 |
| dh | 0.07 | 17.65 | 0.00 | 0.00 | 1.39 | 0.66 | 0.74 |
| eh | 2.36 | 26.36 | 0.00 | 1.55 | 24.46 | 21.46 | 18.08 |
| er | 2.83 | 37.78 | 19.23 | 24.03 | 32.91 | 24.96 | 24.61 |
| ey | 2.26 | 38.61 | 0.56 | 4.98 | 29.36 | 18.92 | 16.19 |
| f | 1.66 | 42.94 | 15.34 | 20.31 | 32.49 | 18.71 | 14.64 |
| g | 0.84 | 51.01 | 5.67 | 8.00 | 14.41 | 5.94 | 5.44 |
| hh | 0.49 | 44.83 | 6.07 | 4.12 | 11.22 | 7.24 | 8.17 |
| ih | 3.97 | 28.00 | 12.34 | 9.90 | 25.00 | 35.26 | 31.93 |
| iy | 4.41 | 64.00 | 0.99 | 13.17 | 64.75 | 45.84 | 39.40 |
| jh | 0.76 | 39.87 | 2.58 | 2.83 | 17.65 | 8.47 | 9.21 |
| k | 4.03 | 66.56 | 22.09 | 27.87 | 41.05 | 21.16 | 21.41 |
| l | 5.75 | 65.68 | 23.02 | 25.87 | 73.81 | 47.24 | 45.38 |
| m | 2.75 | 68.34 | 26.17 | 26.91 | 49.61 | 43.30 | 37.75 |
| n | 5.79 | 64.07 | 27.46 | 40.95 | 69.66 | 59.99 | 58.47 |
| ng | 1.84 | 64.54 | 19.72 | 59.12 | 45.05 | 30.56 | 33.15 |
| ow | 1.62 | 44.91 | 1.63 | 7.79 | 42.60 | 27.20 | 25.21 |
| oy | 0.15 | 13.35 | 0.00 | 0.00 | 3.84 | 2.18 | 1.90 |
| p | 2.04 | 44.08 | 26.51 | 42.66 | 11.82 | 22.08 | 22.56 |
| r | 4.32 | 44.34 | 51.43 | 49.82 | 52.37 | 36.57 | 32.48 |
| s | 9.57 | 60.89 | 2.30 | 50.17 | 66.07 | 53.27 | 55.51 |
| sh | 1.86 | 52.89 | 11.93 | 0.00 | 49.32 | 19.27 | 19.15 |
| t | 5.00 | 33.68 | 21.68 | 22.53 | 19.64 | 20.52 | 20.59 |
| th | 0.26 | 15.07 | 0.00 | 0.00 | 3.52 | 1.68 | 1.67 |
| uh | 0.17 | 7.44 | 0.00 | 0.00 | 0.86 | 0.74 | 0.86 |
| uw | 0.94 | 53.29 | 1.65 | 4.49 | 40.65 | 16.70 | 9.01 |
| v | 1.00 | 35.75 | 9.97 | 3.75 | 14.69 | 15.46 | 15.17 |
| w | 0.81 | 52.31 | 0.36 | 1.08 | 18.41 | 14.66 | 13.01 |
| y | 0.39 | 28.63 | 4.00 | 5.42 | 7.11 | 2.18 | 2.01 |
| z | 4.78 | 45.08 | 17.26 | 19.08 | 41.43 | 29.14 | 29.70 |
| zh | 0.08 | 27.46 | 0.23 | 0.35 | 3.69 | 1.16 | 1.16 |

**Table 2.** Bayesian Kendall's Tau-b between the probability of a phone in the test sample and the combination of Learning (MBL, WH, TD) and Type of Data (Raw and Gaussian).

|                          | Raw   |       |       | Gaussian |       |       |
|--------------------------|-------|-------|-------|----------|-------|-------|
|                          | MBL   | WH    | TD    | MBL      | WH    | TD    |
| **Test-sample Probability** | 0.349 | 0.467 | 0.576 | 0.638    | 0.753 | 0.723 |

To examine whether there is an interaction between Learning Model and Type of Input data we applied Bayesian Quantile Mixed Effect Modeling in the **R** software environment (R Core Team, 2021) using the **brms** package (Bürkner, 2018). Bayesian Quantile Modeling was chosen as a particularly robust and distribution-free alternative to Linear Modeling. We examined the predicted values of the response distribution and their (Bayesian) Credible Intervals (CI) at the median point: Quantile = 0.5 (Schmidtke, Matsuki, & Kuperman, 2017; Tomaschek, Tucker, Fasiolo, & Baayen, 2018, provide more details about criteria for choosing the criterion quantile-point). The final model is summarized in the following formula:

$$\text{SuccessRate} \sim \text{LearningModel} \times \text{TypeOfInput} + (1|\text{Phone})$$

The dependent variable was the success rate, and the two fixed effect factors were learning model (3 levels: MBL, WH, TD) and type of input data (2 levels: Raw, Gaussian). Finally, 40 phone categories represented the random effect factor. The model summary is presented in Figure 3.1.

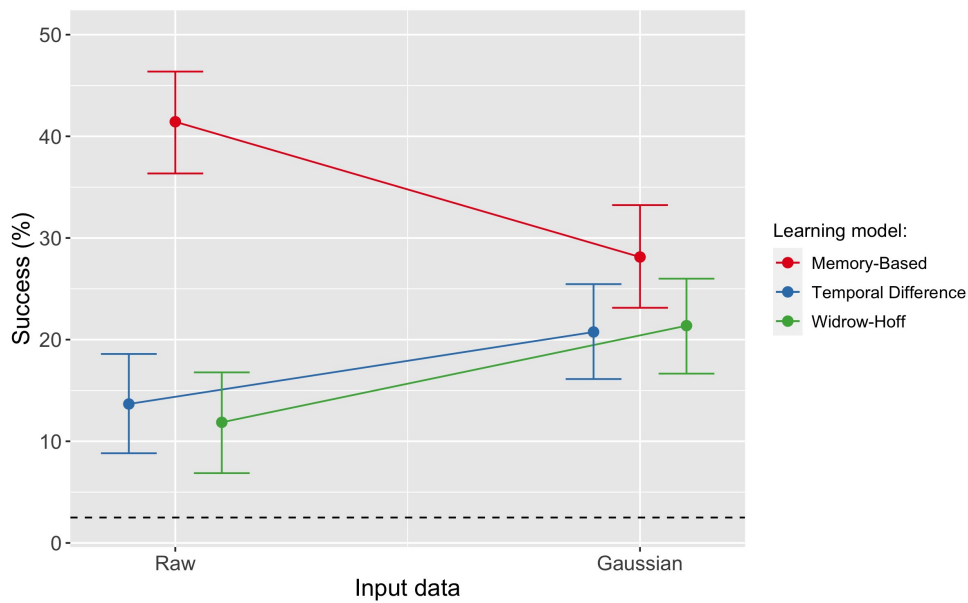

**Figure 1.** Conditional effect of Learning Model and Type of Input on Success Rates with respective 95% Credible Intervals. Black dashed line represents the flat chance level at 2.5%.

If we compare MBL and ECL (jointly WH and TD), first, we see that MBL shows a better overall performance (MBL > ECL: $Estimate = 11.24$ ; $Low\ 90\%\ CI = 8.56$ ; $High\ 90\%\ CI = 13.95$ ; $EvidenceRatio = Inf$ ). That difference, however, attenuates when the Gaussian sample is used (MBL > WH: $Estimate = 6.49$ ; $Low\ 90\%\ CI = 0.49$ ; $High\ 90\%\ CI = 12.32$ ; $EvidenceRatio = 0.04$ ; MBL > TD: $Estimate = 5.88$ ; $Low\ 90\%\ CI = 0.04$; $High\ 90\%\ CI = 11.69$; $EvidenceRatio = 0.05$ ). Thus, we can conclude that MBL performs better with more data (Raw), which is to be expected as this type of dataset provides a very large pool of memorized exemplars. The two ECLs, on the other hand, show improved performance with less data (Gaussian). The improvement is greater for WH than for TD, but that difference is only moderately supported ($Estimate = 2.42$; $Low\ 90\%\ CI = -1.36$; $High\ 90\%\ CI = 6.21$; $EvidenceRatio = 5.86$).

Interestingly, less is more for ECL models: less data gives an overall better performance. Furthermore, with Gaussian data, a 'cleverer', future-sensitive learning model (i.e., Temporal Difference) does not yield better results. It seems that more data leads to overfit (for a detailed discussion, see also Milin et al., 2020). Although it does not cancel out the possibility of overfitting the data, our interpretation of this finding is that the effect is, most likely, yet another consequence of ECL models' sensitivity to the order of trials. As explained above, the Raw data was presented in a particular order, and inasmuch as the ECL models are sensitive to this effect (Milin et al., 2020), the MBL model is immune to it. Further to this, the advantage that TD shows in comparison to WH with Raw data stems from the fact that an immediate reward strategy (as implemented in WH) performs worse than a strategy which considers future rewards when the signal is weak, that is when the data is noisy (for an interesting discussion in the domain of applied robotics see Whelan, Prescott, & Vasilaki, 2021). When that noise is purely Gaussian, however, information about the future does not give much advantage.

Generally speaking, with Gaussian data, the differences between the success rates of the three learning models become insignificant: while MBL becomes less successful, both ECL models benefit. Larger samples with more Gaussian generated data, however, do not improve the performance of ECL further. In fact, larger samples of $N = 1,000; 10,000;$ and $1,866,600$ (the latest to match the size of the Raw data sample) consistently worsened their prediction success (respective averages for Widrow-Hoff learning are: $M_{N=100} = 21.70\%$ ; $M_{N=1K} = 19.49\%$ ; $M_{N=10K} = 14.75\%$; $M_{N=1.8M} = 12.28\%$).

*3.2 Spicing things up: predicting input from a different speaker*

The downward trend of success exhibited by WH learning that is observed as the size of the Gaussian generated sample increases provides indirect support for the assumption that ECL learning rules may indeed be susceptible to overfitting as the data becomes abundant. This sheds light on yet another aspect of the opposition between the two learning principles, MBL and ECL: what 'pleases' the former harms the latter, and vice versa. This difference begs the question of whether fidelity to the data leads to a dead-end once the task becomes even more demanding; that is when, for example, the aim is to predict a new speaker.

For this we use what was learned in the previous step: directly stored pairs of input MFCCs and outcome phones for MBL ($N_{Raw} = 1,866,599$ stored exemplars) and the resulting WH and TD weight matrices (using the same $N_{Raw} = 1,866,599$), to predict phones from MFCCs that were generated by a new female speaker. The dataset consisted of $N_{Raw} = 1,699,643$ test trials from the production of about 27,000 words.

To test the ECL models we used the two weight matrices from training on Raw input, one with WH and another with TD. Then, raw MFCCs from the new speaker were weighted with weights from the two corresponding matrices and then summed to generate the activation for each of the 40 phone outcomes (i.e., the net input), together with their diversity (the absolute length or 1–norm) (cf., Milin, Feldman, et al., 2017). As before, the predicted phone was the one with the strongest support over the competition (the ratio of activation over diversity). To test the MBL model, new MFCC input was 'placed' among the 7-Nearest Neighbors to induce the majority vote, weighted by the system's confidence in that vote (i.e., the size or probability of that majority). The phone winning the majority vote was the predicted phone.

The results from this set of simulations are unanimous: the task of predicting a different speaker is too difficult and all three learning algorithms performed at chance level. The estimated performance (success rates) was given by the following Bayesian Quantile Mixed Effect Model:

$$\text{SuccessRate} \sim \text{LearningModel} + (1|\text{Phone})$$

The resulting estimates for the three learning models are: MBL $Estimate = 2.28$; ($Low\ 95\%\ CI = 1.29$; $High\ 95\%\ CI = 3.36$), TD $Estimate = 1.86$; ($Low\ 95\%\ CI = 0.97$; $High\ 95\%\ CI = 2.93$), WH $Estimate = 1.69$; ($Low\ 95\%\ CI = 0.83$; $High\ 95\%\ CI = 2.79$). In other words, all three learning algorithms failed to generalize to a new speaker.

In an attempt to better understand the results, we generated confusion matrices for each model (see supplementary material). The confusion matrices for the first speaker show that for the MBL model the confusion matrix mirrored what we might expect from a phonetic/perceptual perspective. For example, vowels were most confusable with vowels and /s/ was confusable with /z/. This does not carry over for the TD and WH models where the confusions seem more random. In the confusion matrices for the new speaker the results are much closer to chance for most of the phones. As in the confusion matrices with the original speaker, the MBL model mirrored what might be expected from a perceptual perspective. Generally, across all the confusion matrices there is a bias for silence in the responses. This confusion is likely due to the fact that part of the segment for voiceless stops and affricates consists of silence.

*3.3 Assessing the consistency of learning*

The different and task-dependent profiles of ECL and MBL models pose questions regarding the reliability of different learning approaches. In other words, does their performance remain stable across independent learning sessions, i.e., simulated individual learners' experiences, or is their success or failure rather a matter of serendipity? In principle, learning is a domain general and high-level cognitive capacity and, hence, it ought to show a desirable level of consistency or resilience to random variations and, at the same time, allow a desirable level of systematic differences across individuals. Failure to achieve consistency and predictability across different sets of learning events and/or learners (individuals), would disqualify a learning model as a plausible candidate-mechanism.

It is crucial to note here that this section is aimed at addressing the consistency of learning – what happens if the input samples differ? That is an important question to address since it is known that the samples from which we learn language do differ. It is also known that, with sufficient exposure, learners tend to converge on the same end state. So, to capture the effect that the sample has on what can be learned given the algorithm, an early snapshot is needed, i.e., a snapshot based on small but divergent samples. To have a realistic figure, we settled for the receptive vocabulary of the child at age 2, which is the earliest known sample of sufficient size for modelling purposes.

Given the fact that a two-year old child's receptive vocabulary consists of approximately 300 different words (Hamilton, Plunkett, & Schafer, 2000), we sampled without replacement 5 random samples of 300 words each. These 300-word samples mapped to a somewhat different number of input MFCC and output phone pairs. As before, such input-output pairs were used in the simulations. First, each of the 5 samples of MFCC-phone pairs (i.e., learning events) was replicated 1000 times to simulate solid entrenchment. Next, to keep these learning events as realistic as possible and, at the same time, to prevent overfitting, we added Gaussian noise (with a small deviation) to 50% randomly chosen MFCCs: this reflects minor variations in pronunciation across trials. All five random samples were used to train our three main learning algorithms, both ECLs (Widrow-Hoff and Temporal Difference) and MBL (as before, using Euclidean distances to find the 7-Nearest Neighbors), in five independent learning sessions.

In parallel to this, we generated 5 test samples. Each test sample consisted of 200 words in total. Half of which were 'known' words, i.e., sampled from the corresponding training sample. The other half of words were 'new', not previously encountered. As before, WH and TD learning yielded learning weight matrices, while MBL simply stored all training exemplars as unchanged input-output pairs. Then, the input MFCCs from the test samples were used to predict phone outcomes, as before: (a) for WH and TD the predicted phone was the one with the strongest support against the competition (i.e., the highest ratio of activation over diversity); (b) MBL calculated Euclidean distances to retrieve $k = 7$ nearest neighbors, and to determine their majority vote, weighted by the probability of that vote against all other votes.

This simulation provided success rates for each of the three learning algorithms, separately for each of the five individual learning sessions (generated samples), and for each of the forty phones. As before, we made use of Bayesian Quantile Mixed Effect Modeling as implemented in the **brms** package (Bürkner, 2018) in **R**, and evaluated the model at the median point. This model, too, used the success rate as dependent variable, and the learning model as fixed effect factor (with 3 levels: MBL, WH, TD). In addition to this, it contained two random effects: 40 phones and 5 individual learning sessions:

$$\text{SuccessRate} \sim \text{LearningModel} + (1|\text{Phone}) + (1|\text{Session})$$

Consistent with our previous findings MBL showed a significantly better performance: MBL > ECL: $Estimate = 45.49$; $Low\ 90\%\ CI = 42.04$; $High\ 90\%\ CI = 48.76$; $EvidenceRatio = Inf$. The TD model achieved a marginally better success rate than the WH model: TD > WH: $Estimate = 2.00$; $Low\ 90\%\ CI = -0.61$; $High\ 90\%\ CI = 4.65$; $EvidenceRatio = 8.83$. Strikingly, however, closer inspection of the models performance revealed that in 2 out of 5 sessions MBL's average success was in fact equal to zero. Across the five sessions, the median success rates for MBL were: 72.51, 0.67, 68.00, 71.47, 0.00. Thus, we can conclude that in 40% of cases our simulated 'individuals' did not learn anything at all, while in the other 60% they were luckily successful.

Thus, our follow-up analysis used Median Absolute Deviations (MAD), a robust alternative measure of dispersion, as the target dependent variable. The MAD was calculated for each phone across the five learning sessions. The resulting dataset was submitted to the same Bayesian Quantile Mixed Effect Modeling to test the following model (note that Sessions entered into the calculation of MADs, which is why they could not appear as a random effect):

$$\text{MAD} \sim \text{LearningModel} + (1|\text{Phone})$$

The results confirmed the previously observed significantly higher inconsistency, as indicated by higher MAD, of MBL vs. ECL (MBL > ECL: $Estimate = 5.03$ ; $Low\ 90\%\ CI = 2.37$ ; $High\ 90\%\ CI = 7.96$ ; $EvidenceRatio = 1332.33$ . The two ECL models, however, did not differ significantly among themselves (TD > WH: $Estimate = -0.97$ ; $Low\ 90\%\ CI = -3.68$ ; $High\ 90\%\ CI = 1.83$ ; $EvidenceRatio = 2.55$).

Taken together, these results cast doubt on MBL as viable and plausible mechanism for learning phones from auditory input. On average, MBL performs rather well but it fails the reliability test and, consequently, cannot be considered as a plausible mechanism for learning phones from raw input: too many learners would not learn anything at all. Conversely, while ECL models show poorer average performance, they display higher resilience and consistency across individual learning sessions.

*3.4 Abstracting coherent groups of phones*

From the results presented so far we can be cautiously optimistic about what can be learned from the MFCC input signal. This seems reasonable considering this is the input for most automatic speech recognition systems. There is some systematicity (for details see, for example, Divjak et al., 2021) and that appears to be picked up by learning models. This is, of course, relative to the task at hand, which in this case was predicting a phone from new, raw input MFCCs (the test dataset).

The last challenge for our learning models is to examine whether the learned ECL-weights and MBL-probabilities facilitate the meaningful clustering of phones into traditionally recognized groups. This could be considered indirect evidence for the emergence of (simple linguistic) abstractions. We have assessed the potential of the three learning models to generalize horizontally, i.e., to predict new data from the same speaker and to predict new data from a new speaker. Since traditional classes of phones are well established in the literature, the question is whether our learning models also generalizes vertically. That is, does it yield emerging abstractions of plausible groups of phones which should, at least to a notable extent, match and mirror the existing phonetic groupings of sounds.

To facilitate deliberations on the models' ability to make vertical generalizations and let abstractions emerge, we applied hierarchical cluster analysis using the **pvclust** package (Suzuki & Shimodaira, 2013) in the **R** software environment (R Core Team, 2021) to the respective learning matrices. We used the 'raw' matrices: MBL probabilities and ECL weights.

For simplicity, we focus here on the results for the most and the least successful learning simulations, i.e., WH and MBL using Raw data, and discuss their differences in generating 'natural' groups of phones. We also add the results for TD learning on Raw data, for completeness. The three cluster analyses do not only present the extremes given their prediction successes, but are also the most interesting. Further cluster analyses can be run directly on the weight matrices provided.

All cluster analyses presented here used Euclidean distances and Ward's agglomerative clustering method, with 1,000 bootstrap runs. The results are summarized in Figures 2, 3, and 4. The R package pvclust allows assessing the uncertainty in hierarchical cluster analysis. Using bootstrap resampling, pvclust makes it possible to calculate p-values for each cluster in the hierarchical clustering. The p-value of a cluster ranges between 0 and 1, and indicates how strongly the cluster is supported by data. The package pvclust provides two types of p-values: the AU (Approximately Unbiased) p-value (in red) and the BP (Bootstrap Probability) p-value (in green). The AU p-value, which relies on multiscale bootstrap resampling, is considered a better approximation to an unbiased p-value than the BP value, which is computed by normal bootstrap resampling. The grey values indicate the order of clustering, with smaller values signaling cluster that were formed earlier in the process, and hence contain elements that are considered more similar. The first formed cluster in Figure 2, for example, contains /dh/ and /y/, while the second one contains /oy/ and /uh/. In a third step, the first cluster containing /dh/ and /y/ is joined with /th/.

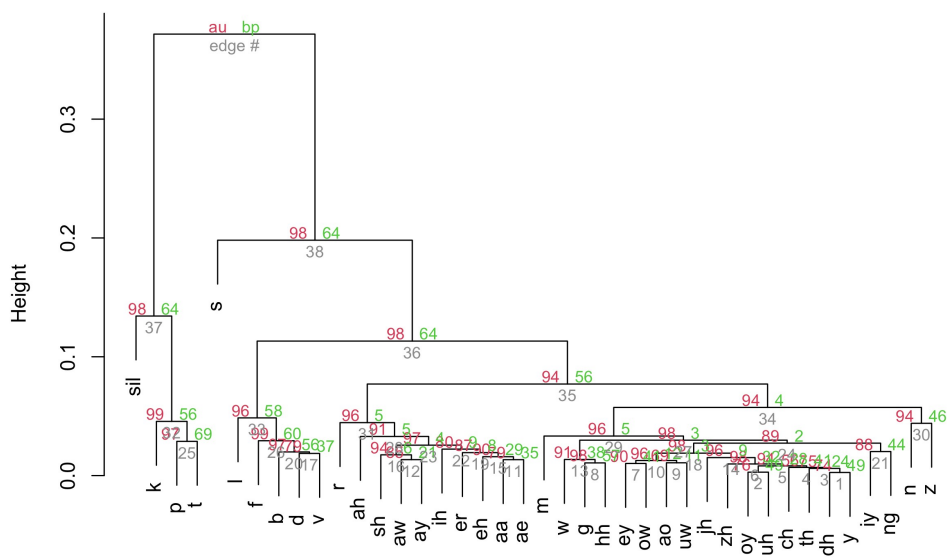

**Figure 2.** Dendrogram of phone clustering using Widrow-Hoff weights from training on Raw data.

The Widrow-Hoff learning weights appear to yield rather intuitive phone groups (Figure 2). Specifically, the left-hand side of the dendrogram reveals small and relatively meaningful clusters of phones. On the right-hand side of the same panel, however, there is a poorly differentiated lump consisting of the 20 remaining phones that do not yet appear to have been learned sufficiently well to form meaningful (i.e., expected) groupings. Nevertheless, some interesting patterns emerge. For example, the voiceless stops cluster together with silence in a later step; considering that a good portion of the voiceless stop is likely silence, this clustering seems fairly logical. The voiceless alveolar fricative /s/ occurs largely by itself and then there is another cluster containing the voiced labial stop /b/, the voiced alveolar stop and lateral /d, l/, and the voiced and voiceless labio-dental fricatives /f, v/. This cluster is largely made up of voice sounds articulated in the front of the vocal tract. The third cluster is largely made up of vowels, a sonorant, and the voiceless palato-alveolar fricative /sh/. This cluster seems to contain most of the vowels along with /r/ and /er/, which have a fairly open vocal tract during articulation. The last cluster contains the remaining 20 phones which seem to be grouped in a mixture of the remaining sounds with no apparent pattern. Note in passing that the WH dendrogram also passes the litmus test of phone recognition, i.e., the distinction between /l/ and /r/. In Figure 2 the /l/ and the /r/ individually belong to clusters that are estimated to replicate on different data.

The results for clustering TD weights are less likely to replicate. Few clusters achieve an AU value of 95 or higher. The clustering is, otherwise, highly similar to the WH dendrogram (compare Figure 2 and Figure 3). There is a later-formed cluster containing the voiceless stops /k, p, t/ on the left-hand side, next to a cluster linking /b, d, m, f, v, g, hh, w/, which includes the voiced stops and some of the bilabial and labio-dental consonants. A third reliable cluster, on the right-hand side, unites 13 phones. These phones are made up of affricates, fricatives and there are a few vowels and a nasal in this group as well. The fourth reliable cluster is so high-level that it links 35 out of the 40 phones together. Note that /s/ and silence are not included in this cluster. Here we also note that the distinction between /l/ and /r/ is less reliable than in the WH analysis.

Figure 4 tells a very different story, or rather: no story at all. The clusters make little sense from a phonetic perspective. Sounds do not seem to cluster together as would be expected and no alternative pattern becomes apparent either. The p-values for the clusters in Figures 2 and 4 reveal some striking differences. Despite some quibbles with the WH phone groups (Figure 2), discussed above, many of the clusters in this solution are rather likely to replicate, and this applies in particular to mid- and high-level clusters (formed later cf. the grey values) on the left-hand side of the diagram. The clusters in the MBL solution, however, show the opposite tendency, where only one cluster (consisting of /oy/ and /uh/) receives an AU p-value equal to or higher than .95. In particular, higher order cluster fare poorly in this respect, with all but one receiving no chance at all of replicating in a different dataset.

Note, also, that the values on the vertical axis are very different for WH/TD and MBL: MBL clusters start forming at value 0.5, by which point the WH and TD cluster solutions have already converged. The chance-like behavior of MBL in clustering what is learned is in line with the highly inconsistent and serendipitous nature of 'learning' exhibited by this particular model.

In summarizing the results of the cluster analysis, two points appear to be of particular interest. First, as pointed out before, learning 'success' does not exist in a vacuum or as a universal notion, but is relative to performance on the task at hand. In that respect, and in the context of the task of predicting new input or indicating higher-order relationships between phones, we can conclude that MBL performs better on predicting unheard input (particularly when the set of exemplars is sufficiently large), but worse on generating meaningful natural groups of phones – i.e., worse on emerging linguistic abstractions. As Nathan (2007) concluded: whatever the role of exemplars might be, they do not make phonemic representations superfluous. We note that the finding that MBL seems to be worse for generating groups of phones seems to contradict the earlier finding that the MBL models show reasonable confusions in the confusion matrices. We believe there are some interesting insights that can be drawn from this potential incompatibility and will be returning to this issue in the general discussion.

The WH rule shows exactly the opposite performance, and is better in generating meaningful natural groups of phones but worse in predicting unheard input. Second, the meaningfulness of the obtained phone groups varies from those that are highly plausible to some that are not sufficiently learned but make a 'primordial soup' of diverse phones. Whether humans experience the same trajectory in the hypothesized acquisition of such phone groups is beyond the scope of this study, however.

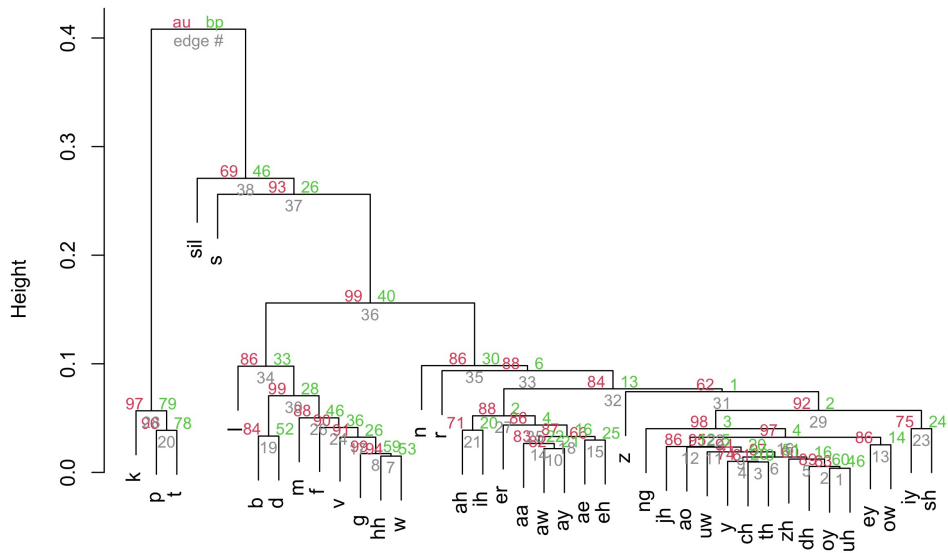

**Figure 3.** Dendrogram of phone clustering using Temporal Difference weights from training on Raw data.

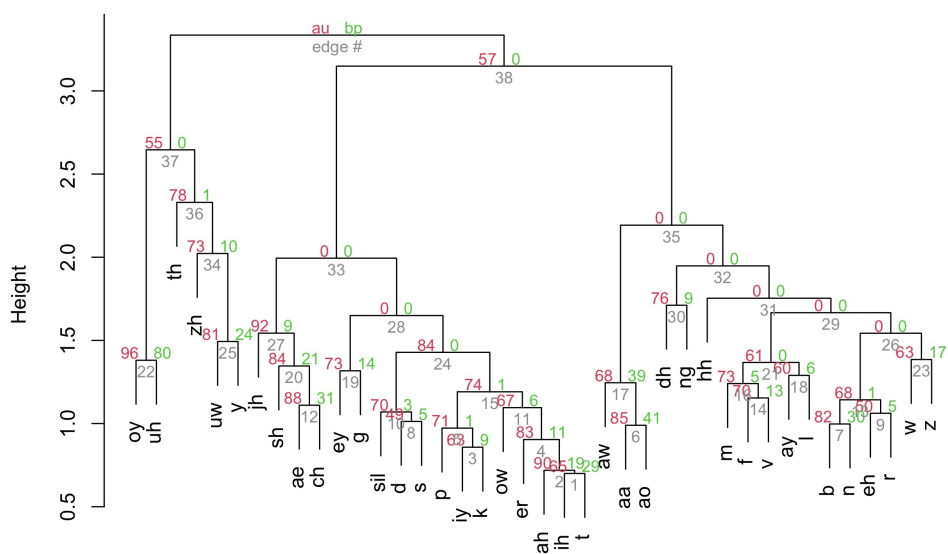

**Figure 4.** Dendrogram of phone clustering using Memory-Based probabilities from training on Raw data.

## 4. General discussion

We set out to test two opposing principles regarding the development of language knowledge in linguistically untrained language users: Memory-Based Learning (MBL) and Error-Correction Learning (ECL). We do not see any algorithm as something the human brain would implement; rather, these algorithms have been devised to capture the operations the brain carries out. We are therefore not claiming that the brain implements the algorithms discussed in the present study (or any other for that matter), only that they can be considered plausible candidates for describing its workings at the higher, conceptual level (i.e., Marr's 1982 computational level of *what the system does*).

MBL relies on a large pool of veridically stored experiences and a mechanism to match new ones with already stored exemplars, focusing on efficiency in storage and matching. ECL filters the input data in such a way that the discrepancy between the system's prediction of the outcome (as made on the basis of available input) and the actual outcome is minimized. Through filtering, those dimensions of the experience that are not useful for reducing the prediction error are removed. What gets stored, thus, changes continuously as the cycles of prediction and confirmation/error-correction alternate. Such an 'evolving unit' cannot be a veridical reflection of an experience, but contains only the most useful essence, distilled from many exposure-events (or usage-events viz. Langacker, 1991).

A process of generalization underlies the abstractions linguists operate with, and we probed whether MBL and/or ECL could give rise to a type of generalized language knowledge that resembles linguistic abstractions. We opted for giving the chosen algorithm a 'hard time' and test their performance under different worst-case scenarios. In that sense, too, our study does not aspire to be biological or psychological plausibility either. The aim is rather to understand what can be learned and what can be generalized from what was learned if the data is 'deprived' rather than (implicitly) enriched. The latter could create a confound in that it becomes difficult to discern whether a particular success (or failure) is due to the data or the algorithm, or perhaps the particular mixture of the two.

Thus, each model was presented with a significant amount of speech (approximately 5 hours) produced by one speaker, created from the MALD stimuli (Tucker et al., 2019). The speech signals were transformed into MFCC representations and then paired with the 1-hot encoded outcome phone. Against such data we pitted three learning rules, a memory greedy one (MBL), a short-sighted one (WH), and one that explicitly takes into account future data (TD). These three rules, again, map onto the principles of Memory-Based vs. Error-Correction Learning (MBL vs. ECL: WH and TD). We put both types of learning algorithms through four tests that probe the quality of their learning by assessing the models' ability to predict data that is very similar to the data they were trained on (same speaker, new words) as well as data that is rather different from the data they were trained on (different speaker, same/different words). We also assessed the consistency or stability of what the models have learned and, finally, their ability to give rise to abstract categories. As expected, both types of models fare differently with regard to these tests.

### 4.1 The learnability of abstractions

While MBL outperforms ECL in predicting new input from the same speaker, this is only the case for Raw data. With Gaussian data, the learning models do not differ significantly in their performance. Interestingly, for ECL less is more: training WH and TD on fewer data points with Gaussian noise yields better prediction results than training WH on Raw and big data. This indicates that WH may in fact overfit in case of larger datasets, essentially overcommitting to the already processed data. Another plausible interpretation of these results is that ECLs, when learning from Raw data, are affected by the order of trials. Yet, neither an order effect nor overfitting to input data are necessarily unnatural or negative as they represent known phenomena in, e.g., language acquisition more generally (see, for example, Arnon & Ramscar, 2012 or Ramscar & Yarlett, 2007) as well as in in L1 and L2 learning of phonetic input specifically.

Infants, at least in Western cultures, typically have one primary caregiver, which makes it likely that they, too, *overfit their mental models* to the input of that particular caregiver. There is evidence from learning in other sensory domains, i.e., vision, that casts this approach in a positive light: infants use their limited world to become experts at a few things first, such as their caregiver's face(s). The same, very few faces are in the infant's visual field frequently and persistently and these extended, repeated exposures have been argued to form the starting point for all visual learning (Jayaraman & Smith, 2019).

L2 learners are also know to overfit their input data, and this tendency has been found to hinder their progress (for a review, see Barriuso & Hayes-Harb, 2018). To remedy the problem, researchers have experimented with high versus low variability phonetic training. High variability conditions include input with multiple talkers and contexts and is hypothesized to support the formation of robust sound categories and boost the generalizability to new talkers. While it has been shown that adult L2 learners benefit from multiple talkers and contexts (for the initial report, see Lively, Logan, & Pisoni, 1993), child L2 learners do not appear to benefit more from high variability input (Barriuso & Hayes-Harb, 2018; Brekelmans, Evans, & Wonnacott, 2020; Giannakopoulou, Brown, Clayards, & Wonnacott, 2017).

However, while overfitting may be an advantage in some learning situations it may cause challenges in others. The use of a single speaker may also introduce a challenge to the model, not dissimilar to the long-known speaker normalization problem in speech perception (e.g., Ladefoged & Broadbent, 1957). Speaker normalization is the process of a listener adapting to sounds (typically vowels) produced by different speakers. This is due to the fact that different speakers have individual characteristics and the acoustic characteristics of a vowel produced by a male speaker may be similar to the acoustic characteristics of an entirely different vowel produced by a female speaker (e.g., Barreda & Nearey, 2012). Speaker normalization describes research in speech perception investigating how listeners adapt to individual speakers. In our modeling, the learner has not been given an opportunity to acquire knowledge about other speakers and as a result it may be that it has not learned to create categories that are speaker independent.

At some point, however, learners are exposed to input produced by another speaker than their primary care-giver or teacher. We simulated this situation by exposing our models, trained on input from one speaker, to input produced by a novel speaker. This time, unfortunately, all learning models performed at chance level, showing that they were unable to generalize what they had learned to a new speaker. This did not come as a complete surprise, however, as the challenge was deliberately extreme, not allowing for any prior experience with that speaker at all. Incrementally adding input from different speakers into the mix will be informative about how long and to what extent the model can or must keep on learning to be able to generalize to new, previously unheard speakers. This, too, remains beyond the scope of the present exploratory study.

The idea of individual differences in early learning was addressed under a set of simplifying assumptions, for the worst-case scenario: we simulated 5 child-like learners that each experienced plenty of minimally different examples of a limited number of words, spoken by one and the same speaker. When tested on seen and unseen words, MBL appeared to outperform ECL, but closer inspection of the model revealed that this advantage applied to a non-existent average only: 2 out of the 5 learners who took a MBL approach had learned nothing at all. This, then, casts doubts on the cognitive plausibility of this particular learning approach, which demonstrates as desirable at the population level but fails many individuals. Contrary to MBL's performance, both ECL models showed lower success on average, yet they are the more reliable performance across simulated 'individuals'.

In a last step, we tested to what extent weights (from the two ECLs) and probabilities (from MBL), which we understand as summative proxies of learning, facilitate the meaningful clustering of phones into traditionally acknowledged groups of phones. Even though ECL was trained on its least favorite type of data (Raw) it performed reasonably well: it gave rise to a dendrogram that allocated at least half of the phone inventory to groups that resemble their traditional phonetic classification. Furthermore, the groups it identified were estimated to have a high chance of being detected in new, independent datasets. MBL, on the other hand, performed exceptionally poorly, even though it was trained on its favorite type of data (again, Raw). The absence of any recognizable systematicity in the MBL-based clustering, and the estimated lack of replicability of the clusters, is striking indeed.

A word of caution is in place regarding the apparent inconsistency of the results of both types of algorithms, MBL and ECL, where the former was better in predicting new phones for the same speaker, and the latter was better in creating meaningful groups of phones. This observation constitutes the focal point of our final subsection.

*4.2 Implications for a theoretical discussion of abstractions*

Generalization is one of the most basic attributes of the broad notion of learning. To quote one of the pioneers of Artificial Neural Networks (ANNs), intelligent Signal Processing and Machine Learning: "if generalization is not needed, we can simply store the associations in a look-up table, and will have little need for a neural network", where the "inability to realize all functions is in a sense a strength [...] because it [...] improves its ability to generalize" to new and unseen patterns (Widrow & Lehr, 1990, p. 1422). From this, by extension, we are in position to draw several conclusions. First, the set task appeared to require very little, if any, generalization. We allowed for learning from abundant data of the same speaker, where all possible outcomes (phones) were experienced (stored or filtered) many times. Second, the amount of training data, indeed, created an excellent "look-up table" for MBL to find the right neighborhoods for a given exemplar. Three, that same training data most likely had two strong effects on ECL learning: trial-order and overfit, and they both led to poorer performance on the task of predicting a phone from the same speaker, the task that did not require a capacity to generalize. Fourth, when the task was designed to rely on generalizations and to abstract away from individual phones to their higher-order classes or types, the ECL models did, actually, much better. MBL, conversely, failed to 'see the forest for the trees'. Again, with regard to apparent inconsistency of results – better prediction of MBL vs. better base of cluster formation of ELC, it is important to keep in mind that the tasks were different given that the former tested the accuracy of identification while the latter tested the potency to generalize, if generalization is operationalized as clustering.

At the very least, our findings show that we should not throw the abstractions-baby out with the theoretical bathwater. Instead, the way in which we approach the discussion of abstractions should change. Up until now, the amount of theoretical speculation about the existence of abstractions exceeds the amount of empirical work. This is specifically true for attempts to model how such abstractions might emerge from exposure to input.

For example, Ramscar and Port (2016) discuss the impossibility of a discrete inventory of phonemes and claim that "none of the traditional discrete units of language that are thought to be the compositional parts of language – phones or phonemes, syllables, morphemes, and words – can be identified unambiguously enough to serve as the theoretical bedrock on which to build a successful theory of spoken language composition and interpretation. Surely, if these units were essential components of speech perception, they would not be so uncertainly identified" (p.63). Our results show that 'there must be more to that': the findings are not unanimous but speak for much more than no-thing and much less than every-thing. We have shown that at least part of the phone inventory can be reliably identified from input by algorithms applied here – ECL and MBL. At the same time, not all phones appear to be equally learnable from exposure, at least not from exposure to one speaker. For any balanced diet, the proof is in the pudding, but neither in the know-it-all chef or the list of ingredients.

Maybe this should not come as a surprise: after all, many if not most abstractions came into existence as handy theoretical or descriptive shortcuts, not as cognitively realistic reflections of (individual) users' language knowledge (see, for discussion, Divjak, 2015b; Milin et al., 2016). Cognitively speaking, the phoneme inventory that is traditionally assumed for English is possibly either too specific or too coarse in places, in particular when dealing with the idiosyncrasies of one speaker. Schatz et al., 2021 likewise reported finding a different type of units, as the ones their model learned were too brief and too variable acoustically to correspond to the traditional phonetic categories. They, too, took the view that such a result should not automatically be taken to challenge the learning mechanism, but rather to challenge what is expected to be learned, e.g., phonetic categories. Or maybe we should consider that not all phones appear to be equally learnable from input alone and production data should be provided to the model; emergentist approaches allow production constraints to affect perception (Donegan, 2015, p. 45). Perhaps, also, the idiosyncrasies of our one speaker show that more diverse training is necessary to learn other groups of phone-like units.

Shortcomings aside, we have shown that ECL learning methods can learn abstractions. Continuous-time learning, as represented by TD, does not necessarily improve learning as measured by overall prediction accuracy, but looking into the future helps if the data overlaps temporally. Furthermore, the argument for or against the relevance of units and abstractions is, as we said, void if not pitted against a particular task that may or may not require such units to be discerned and/or categorized (this point was demonstrated by successful models of categorization, such as Love et al., 2004; Nosofsky, 1986). In this sense, ECL offers the more versatile type of learning in that it can handle tasks at the level of exemplars and of abstractions, as demonstrated by better performance on average, across the four tasks. It also confirms claims that adults seem to disregard differences that are not phonemic rather than actually losing the ability to perceive them (Donegan, 2015, p. 43).

The empirical exploration of whether abstractions are learnable should precede any decisions about which theoretical framework to promote. The Zeitgeist, however, seems to favor a reductionism that leads to rather extreme positions on the role of exemplars and the status of (abstract) units (cf., Ambridge, 2020; Ramscar, 2019). Yet, as we know from the past, reductionism of any kind provokes rather than inspires (recall, for example, the Behaviorist take on Language, and the Generativist reaction to it, see, Chomsky, 1959; Skinner, 1957; and for a later re-evaluation of these contributions, see, among others, MacCorquodale, 1970; Palmer, 2006; Swiggers, 1995), where both provocateur and provokee lose in the long run. Such theoretical deliberations only lead to 'tunnel vision' on what constitutes evidence and how supporting data ought to be collected. After all, we should know better: "all models are wrong" (Box, 1976, p. 792), and those that occupy extreme positions often go hand-in-hand with simplifying and even simplistic viewpoints, as their potential usefulness gets spent on promotion and debasement rather than on productive (self-)refutability.

**Supplementary Materials**

Confusion matrices are readily available on OSF repository at: https://osf.io/4xnwy/.

**Acknowledgments**

This work was funded by a Leverhulme Trust award (RL-2016-001) which funded PM and DD and by the Social Sciences and Humanities Research Council of Canada (4352014-0678) which funded BVT. We are grateful to Ben Ambridge for a fascinating seminar which sparked our interest in this topic, and to members of the Out of our Minds team and the Alberta Phonetics Laboratory for sharing their thoughts. We thank Danielle Matthews, Gerardo Ortega and Eleni Vasilaki for useful discussions and/or pointers to the literature.

## References

Ambridge, B. (2020). Against stored abstractions: A radical exemplar model of language acquisition. *First Language, 40*(5-6), 509-559. doi:10.1177/0142723719869731

Ambridge, B., & Lieven, E. V. (2011). *Child language acquisition: Contrasting theoretical approaches*: Cambridge University Press.

Anderson, J. R. (2000). *Learning and memory: An integrated approach*: John Wiley & Sons Inc.

Anderson, N. D., Holmes, E. W., Dell, G. S., & Middleton, E. L. (2019). Reversal shift in phonotactic learning during language production: Evidence for incremental learning. *Journal of Memory and Language, 106*, 135-149.

Appelbaum, I. (1996). *The lack of invariance problem and the goal of speech perception*. Paper presented at the Proceeding of Fourth International Conference on Spoken Language Processing. ICSLP '96.

Arnold, D., & Tomaschek, F. (2016). The Karl Eberhards Corpus of spontaneously spoken Southern German in dialogues-audio and articulatory recordings.

Arnold, D., Tomaschek, F., Sering, K., Lopez, F., & Baayen, R. H. (2017). Words from spontaneous conversational speech can be recognized with human-like accuracy by an error-driven learning algorithm that discriminates between meanings straight from smart acoustic features, bypassing the phoneme as recognition unit. *Plos One, 12*(4), e0174623.

Arnon, I., & Ramscar, M. (2012). Granularity and the acquisition of grammatical gender: How order-of-acquisition affects what gets learned. *Cognition, 122*(3), 292-305.

Baayen, R. H., Chuang, Y.-Y., Shafaei-Bajestan, E., & Blevins, J. P. (2019). The discriminative lexicon: A unified computational model for the lexicon and lexical processing in comprehension and production grounded not in (de) composition but in linear discriminative learning. *Complexity, 2019*.

Baayen, R. H., Milin, P., Đurđević, D. F., Hendrix, P., & Marelli, M. (2011). An amorphous model for morphological processing in visual comprehension based on naive discriminative learning. *Psychological Review, 118*(3), 438.

Barreda, S., & Nearey, T. M. (2012). The association between speaker-dependent formant space estimates and perceived vowel quality. *Canadian Acoustics, 40*(3), 12-13.

Barriuso, T. A., & Hayes-Harb, R. (2018). High Variability Phonetic Training as a Bridge from Research to Practice. *CATESOL Journal, 30*(1), 177-194.

Beck, J. M., Ma, W. J., Kiani, R., Hanks, T., Churchland, A. K., Roitman, J., Shadlen, M. N., Latham, P. E., & Pouget, A. (2008). Probabilistic population codes for Bayesian decision making. *Neuron, 60*(6), 1142-1152.

Blevins, J. P. (2016). *Word and paradigm morphology / James P. Blevins* (First edition. ed.): Oxford : Oxford University Press, 2016.

Bod, R. (1998). Beyond grammar: An experience-based theory of language.

Boerlin, M., Machens, C. K., & Denève, S. (2013). Predictive coding of dynamical variables in balanced spiking networks. *PLoS computational biology, 9*(11), e1003258.

Bouton, M. E. (2007). *Learning and Behavior: A Contemporary Synthesis*. Sunderland, MA: Sinauer Associates.

Box, G. E. (1976). Science and statistics. *Journal of the American Statistical Association, 71*(356), 791-799.

Brekelmans, G., Evans, B., & Wonnacott, E. (2020). *No evidence of a high variability benefit in phonetic vowel training for children.* Paper presented at the Book of Abstracts: 2nd Workshop on Speech Perception and Production across the Lifespan (SPPL2020). 2nd Workshop on Speech Perception and Production across the Lifespan (SPPL2020), London, UK.

Buchwald, A., & Miozzo, M. (2011). Finding Levels of Abstraction in Speech Production. *Psychological Science, 22*(9), 1113-1119. doi:10.1177/0956797611417723

Buesing, L., Bill, J., Nessler, B., & Maass, W. (2011). Neural dynamics as sampling: a model for stochastic computation in recurrent networks of spiking neurons. *PLoS computational biology, 7*(11), e1002211.

Bürkner, P.-C. (2018). Advanced Bayesian Multilevel Modeling with the R Package brms. *The R Journal, 10*(1). doi:10.32614/rj-2018-017

Bush, R. R., & Mosteller, F. (1955). Stochastic models for learning / Robert R. Bush, Frederick Mosteller. In. New York : London: New York : Wiley London : Chapman & Hall.

Bybee, J. L. (2013). Usage-based theory and exemplar representations of constructions. In T. Hoffmann & G. Trousdale (Eds.), *The Oxford handbook of construction grammar* (pp. 49-69). Oxford: Oxford University Press.

Chomsky, N. (1959). A Review of BF Skinner's Verbal Behavior. *Language, 35*(1), 26-58.

Chuang, Y.-Y., & Baayen, R. H. (2021). Discriminative learning and the lexicon: NDL and LDL. In: Oxford Research Encyclopedia of Linguistics (in press). Oxford University Press.

Cover, T., & Hart, P. (1967). Nearest neighbor pattern classification. *IEEE transactions on information theory, 13*(1), 21-27.

Daelemans, W., & Van den Bosch, A. (2005). *Memory-based language processing*. Cambridge: Cambridge University Press.

Divjak, D. (2015a). Exploring the grammar of perception: A case study using data from Russian. *Functions of Language, 22*(1), 44-68.

Divjak, D. (2015b). Four challenges for usage-based linguistics. In *Change of Paradigms–New Paradoxes* (pp. 297-310): De Gruyter Mouton.

Divjak, D. (2019). *Frequency in language : memory, attention and learning / Dagmar Divjak*: Cambridge : Cambridge University Press, 2019.

Divjak, D., Milin, P., Ez-zizi, A., Józefowski, J., & Adam, C. (2021). What is learned from exposure: an error-driven approach to productivity in language. *Language, Cognition and Neuroscience, 36*(1), 60-83.

Divjak, D., Szymor, N., & Socha-Michalik, A. (2015). Less is more: possibility and necessity as centres of gravity in a usage-based classification of core modals in Polish. *Russian Linguistics, 39*(3), 327-349.

Donegan, P. (2015). The Emergence of Phonological Representation. In *The Handbook of Language Emergence* (pp. 33-52).

Ellis, N. C. (2006a). Language acquisition as rational contingency learning. *Applied linguistics, 27*(1), 1-24.

Ellis, N. C. (2006b). Selective attention and transfer phenomena in L2 acquisition: Contingency, cue competition, salience, interference, overshadowing, blocking, and perceptual learning. *Applied linguistics, 27*(2), 164-194.

Ellis, N. C. (2016). Salience, cognition, language complexity, and complex adaptive systems. *Studies in Second Language Acquisition, 38*(2), 341-351.

Ferro, M., Marzi, C., & Pirrelli, V. (2011). A self-organizing model of word storage and processing: implications for morphology learning. *Lingue e linguaggio, 10*(2), 209-226.

Fitz, H., & Chang, F. (2019). Language ERPs reflect learning through prediction error propagation. *Cognitive Psychology, 111*, 15-52.

Fromkin, V. A. (1971). The Non-Anomalous Nature of Anomalous Utterances. *Language, 47*(1). doi:10.2307/412187

Gershman, S. J. (2015). A unifying probabilistic view of associative learning. *PLoS computational biology, 11*(11), e1004567.

Giannakopoulou, A., Brown, H., Clayards, M., & Wonnacott, E. (2017). High or low? Comparing high and low-variability phonetic training in adult and child second language learners. *PeerJ, 5*. doi:10.7717/peerj.3209

Gillies, J. (2018). *CERN and the Higgs Boson: The Global Quest for the Building Blocks of Reality*. London: Icon Books.

Goldberg, A. E. (2006). *Constructions at work: The nature of generalization in language*: Oxford University Press on Demand.

Goldinger, S. D. (1998). Echoes of echoes? An episodic theory of lexical access. *Psychological Review, 105*(2), 251-279.

Goldinger, S. D., & Azuma, T. (2003). Puzzle-solving science: the quixotic quest for units in speech perception. *Journal of Phonetics, 31*(3-4), 305-320. doi:10.1016/s0095-4470(03)00030-5

Graves, A., & Schmidhuber, J. (2005). Framewise phoneme classification with bidirectional LSTM and other neural network architectures. *Neural networks, 18*(5-6), 602-610.

Grossberg, S. (2013). Adaptive Resonance Theory: How a brain learns to consciously attend, learn, and recognize a changing world. *Neural networks, 37*, 1-47.

Haefner, R. M., Berkes, P., & Fiser, J. (2016). Perceptual decision-making as probabilistic inference by neural sampling. *Neuron, 90*(3), 649-660.

Hamilton, A., Plunkett, K. I. M., & Schafer, G. (2000). Infant vocabulary development assessed with a British communicative development inventory. *Journal of Child Language, 27*(3), 689-705. doi:10.1017/s0305000900004414

Hay, J., & Bresnan, J. (2006). Spoken syntax: The phonetics of giving a hand in New Zealand English. *The Linguistic Review, 23*(3). doi:10.1515/tlr.2006.013

Haykin, S. S. (1999). Neural networks : a comprehensive foundation / Simon Haykin. In (2nd ed.). London: London : Prentice Hall.

Hebb, D. O. (1949). *The organization of behavior*. New York: Wiley.

Hull, C. L. (1943). Principles of behavior: An introduction to behavior theory.

Jääskeläinen, I. P., Hautamäki, M., Näätänen, R., & Ilmoniemi, R. J. (1999). Temporal span of human echoic memory and mismatch negativity: revisited. *Neuroreport, 10*(6), 1305-1308.

Jayaraman, S., & Smith, L. B. (2019). Faces in early visual environments are persistent not just frequent. *Vision research, 157*, 213-221.

Jebara, T. (2012). *Machine learning: discriminative and generative* (Vol. 755): Springer Science & Business Media.

Johnson, K. (1997). The auditory/perceptual basis for speech segmentation.

Julià, P. (1983). *Explanatory models in linguistics : a behavioral perspective*. Princeton: Princeton University Press.

Kamin, L. J. (1969). Predictability, surprise, attention, and conditioning. In B. Campbell & R. Church (Eds.), *Punishment and aversive behaviour* (pp. 279-296). New York: Appleton-Century-Crofts.

Kokkola, N. H., Mondragón, E., & Alonso, E. (2019). A double error dynamic asymptote model of associative learning. *Psychological Review, 126*(4), 506.

Ladefoged, P., & Broadbent, D. E. (1957). Information Conveyed by Vowels. *The Journal of the Acoustical Society of America, 29*(1), 98-104. doi:10.1121/1.1908694

Langacker, R. W. (1987). *Foundations of cognitive grammar: Theoretical prerequisites* (Vol. 1): Stanford university press.

Langacker, R. W. (1991). *Foundations of Cognitive Grammar: Descriptive application* (Vol. 2). Stanford: Stanford university press.

Langacker, R. W. (2017). Entrenchment in Cognitive Grammar. In H.-J. Schmid (Ed.), *Entrenchment and the psychology of language learning: how we reorganize and adapt linguistic knowledge* (pp. 39-56). Berlin: De Gruyter Mouton & APA.

Liberman, A. M., Cooper, F. S., Shankweiler, D. P., & Studdert-Kennedy, M. (1967). Perception of the speech code. *Psychological Review, 74*(6), 431.

Lively, S. E., Logan, J. S., & Pisoni, D. B. (1993). Training Japanese listeners to identify English /r/ and /l/. II: The role of phonetic environment and talker variability in learning new perceptual categories. *The Journal of the Acoustical Society of America, 94*(3), 1242-1255. doi:10.1121/1.408177

Love, B. C., Medin, D. L., & Gureckis, T. M. (2004). SUSTAIN: A Network Model of Category Learning. *Psychological Review, 111*(2), 309-332. doi:10.1037/0033-295x.111.2.309

MacCorquodale, K. (1970). On Chomsky's review of Skinner's Verbal behavior. *Journal of the experimental analysis of behavior, 13*(1), 83.

Marr, D. (1982). *Vision: A computational investigation into the human representation and processing of visual information*. San Francisco: Freeman & Co.

Marslen-Wilson, W., & Warren, P. (1994). Levels of perceptual representation and process in lexical access: Words, phonemes, and features. *Psychological Review, 101*(4), 653-675. doi:10.1037/0033-295x.101.4.653

McClelland, J. L., & Rumelhart, D. E. (1989). *Explorations in parallel distributed processing: A handbook of models, programs, and exercises*: MIT press.

Mendel, J., & McLaren, R. (1970). 8 reinforcement-learning control and pattern recognition systems. In *Mathematics in science and engineering* (Vol. 66, pp. 287-318): Elsevier.

Mielke, J., Baker, A., & Archangeli, D. (2016). Individual-level contact limits phonological complexity: Evidence from bunched and retroflex /ɹ/. *Language, 92*(1), 101-140. doi:10.1353/lan.2016.0019

Milin, P., Divjak, D., & Baayen, R. H. (2017). A learning perspective on individual differences in skilled reading: Exploring and exploiting orthographic and semantic discrimination cues. *Journal of Experimental Psychology: Learning, Memory, and Cognition, 43*(11), 1730.

Milin, P., Divjak, D., Dimitrijević, S., & Baayen, R. H. (2016). Towards cognitively plausible data science in language research. *Cognitive Linguistics, 27*(4), 507-526. doi:10.1515/cog-2016-0055

Milin, P., Feldman, L. B., Ramscar, M., Hendrix, P., & Baayen, R. H. (2017). Discrimination in lexical decision. *Plos One, 12*(2), e0171935.

Milin, P., Keuleers, E., & Đurđević, D. (2011). Allomorphic responses in Serbian pseudo-nouns as a result of analogical learning. *Acta Linguistica Hungarica, 58*(1), 65-84.

Milin, P., Madabushi, H. T., Croucher, M., & Divjak, D. (2020). Keeping it simple: Implementation and performance of the proto-principle of adaptation and learning in the language sciences. *arXiv preprint arXiv:2003.03813*.

Minsky, M., & Papert, S. (1969). *Perceptrons: An Introduction to Computational Geometry*. Cambridge, MA: The MIT Press.

Nenadić, F. (2020). *Computational Modelling of Spoken Word Recognition in the Auditory Lexical Decision Task.* (PhD), University of Alberta, Edmonton, Canada.

Nesdale, A. R., Herriman, M. L., & Tunmer, W. E. (1984). Phonological awareness in children. In *Metalinguistic awareness in children: Theory, research, and implications* (pp. 56-72). Berlin: Springer.

Nixon, J. S., & Tomaschek, F. (2020). *Learning from the acoustic signal: Error-driven learning of low-level acoustics discriminates vowel and consonant pairs.* Paper presented at the CogSci.

Nosofsky, R. M. (1986). Attention, similarity, and the identification–categorization relationship. *Journal of experimental psychology: General, 115*(1), 39.

Nosofsky, R. M. (1988). Similarity, frequency and category representation. *Journal of Experimental Psychology: Learning, Memory, and Cognition, 14*, 54-65.

Osgood, C. E., Sebeok, T. A., Gardner, J. W., Carroll, J. B., Newmark, L. D., Ervin, S. M., Saporta, S., Greenberg, J. H., Walker, D. E., & Jenkins, J. J. (1954). Psycholinguistics: a survey of theory and research problems. *The Journal of Abnormal and Social Psychology, 49*(4p2), i.

Palmer, D. C. (2006). On Chomsky's appraisal of Skinner's Verbal Behavior: A half century of misunderstanding. *The Behavior Analyst, 29*(2), 253-267.

Pierrehumbert, J. (2001). Lenition and contrast. *Frequency and the emergence of linguistic structure, 45*, 137.

Port, R. F. (2007). How are words stored in memory? Beyond phones and phonemes. *New Ideas in Psychology, 25*(2), 143-170. doi:10.1016/j.newideapsych.2007.02.001

Port, R. F. (2010). Language as a Social Institution: Why Phonemes and Words Do Not Live in the Brain. *Ecological Psychology, 22*(4), 304-326. doi:10.1080/10407413.2010.517122

Port, R. F., & Leary, A. P. (2005). Against formal phonology. *Language, 81*(4), 927-964.

R Core Team. (2021). R: A language and environment for statistical computing (Version 4.1.1). Vienna, Austria: R Foundation for Statistical Computing. Retrieved from https://www.R-project.org/

Ramscar, M. (2019). Source codes in human communication. *arXiv preprint arXiv:1904.03991*.

Ramscar, M., Dye, M., & McCauley, S. M. (2013). Error and expectation in language learning: The curious absence of" mouses" in adult speech. *Language*, 760-793.

Ramscar, M., & Port, R. F. (2016). How spoken languages work in the absence of an inventory of discrete units. *Language Sciences, 53*, 58-74.

Ramscar, M., & Yarlett, D. (2007). Linguistic self-correction in the absence of feedback: A new approach to the logical problem of language acquisition. *Cognitive science, 31*(6), 927-960.

Ramscar, M., Yarlett, D., Dye, M., Denny, K., & Thorpe, K. (2010). The effects of feature-label-order and their implications for symbolic learning. *Cognitive science, 34*(6), 909-957.

Rescorla, R. (2008). Rescorla-Wagner model. *Scholarpedia, 3*(3), 2237.

Rescorla, R. A. (1988). Pavlovian conditioning: It's not what you think it is. *American psychologist, 43*(3), 151-160.

Rescorla, R. A., & Wagner, R. A. (1972). A theory of Pavlovian conditioning: variations in the effectiveness of reinforcement and non-reinforcement. In H. Black & W. F. Proksay (Eds.), *Classical conditioning II* (pp. 64-99). New York, NY: Appleton-Century-Crofts.

Romain, L., Ez-zizi, A., Milin, P., & Divjak, D. (2022). What makes the past perfect and the future progressive?: Experiential coordinates for a learnable, context-based model of tense and aspect. *Cognitive Linguistics*.

Rosch, E. (1975). Cognitive representation of semantic categories. *Journal of Experimental Psychology 104*(3 ), 192-233.

Rosenblatt, F. (1958). The perceptron: a probabilistic model for information storage and organization in the brain. *Psychological Review, 65*(6), 386-408.

Sapir, E. (1921). *Language: An introduction to the study of speech*: Harcourt, Brace.

Savin, H. B., & Bever, T. G. (1970). The nonperceptual reality of the phoneme. *Journal of Verbal Learning and Verbal Behavior, 9*(3), 295-302.

Schatz, T., Feldman, N. H., Goldwater, S., Cao, X.-N., & Dupoux, E. (2021). Early phonetic learning without phonetic categories: Insights from large-scale simulations on realistic input. *Proceedings of the National Academy of Sciences, 118*(7).

Schmidtke, D., Matsuki, K., & Kuperman, V. (2017). Surviving blind decomposition: A distributional analysis of the time-course of complex word recognition. *Journal of Experimental Psychology: Learning, Memory, and Cognition, 43*(11), 1793.

Shafaei-Bajestan, E., Moradipour-Tari, M., Uhrig, P., & Baayen, R. H. (2021). LDL-AURIS: a computational model, grounded in error-driven learning, for the comprehension of single spoken words. *Language, Cognition and Neuroscience*, 1-28. doi:10.1080/23273798.2021.1954207

Shankweiler, D., & Fowler, C. A. (2015). Seeking a reading machine for the blind and discovering the speech code. *History of Psychology, 18*(1), 78-99. doi:10.1037/a0038299

Shoup, J. E. (1980). Phonological aspects of speech recognition. *Trends in speech recognition*, 125-138.

Skinner, B. F. (1957). *Verbal behavior*: New York: Appleton-Century-Crofts.

Sutton, R. S., & Barto, A. G. (1987). *A temporal-difference model of classical conditioning.* Paper presented at the Proceedings of the ninth annual conference of the cognitive science society.

Sutton, R. S., & Barto, A. G. (1990). Time-derivative models of Pavlovian reinforcement.

Suzuki, R., & Shimodaira, H. (2013). Hierarchical clustering with P-values via multiscale bootstrap resampling. *R package*.

Swiggers, P. (1995). How Chomsky skinned Quine, or what 'verbal behavior'can do. *Language Sciences, 17*(1), 1-18.

Taylor, J. R. (1995). *Linguistic categorization: Prototypes in linguistic theory*: Clarendon Press.

Tolman, E. C. (1932). *Purposive behavior in animals and men*: Univ of California Press.

Tomaschek, F., Tucker, B. V., Fasiolo, M., & Baayen, R. H. (2018). Practice makes perfect: The consequences of lexical proficiency for articulation. *Linguistics Vanguard, 4*(s2).

Trubetzkoy, N. S. (1969). Principles of phonology.

Tucker, B. V., Brenner, D., Danielson, D. K., Kelley, M. C., Nenadić, F., & Sims, M. (2019). The massive auditory lexical decision (MALD) database. *Behavior research methods, 51*(3), 1187-1204.

Tunmer, W. E., & Rohl, M. (1991). Phonological awareness and reading acquisition. In *Phonological awareness in reading: The evolution of current perspectives* (pp. 1-30). Berlin: Springer.

Vértes, E., & Sahani, M. (2019). A neurally plausible model learns successor representations in partially observable environments. *Advances in*

*Neural Information Processing Systems, 32*, 13714-13724.

Whelan, M. T., Prescott, T. J., & Vasilaki, E. (2021). A robotic model of hippocampal reverse replay for reinforcement learning. *arXiv preprint arXiv:2102.11914*.

Widrow, B., & Hoff, M. E. (1960). *Adaptive switching circuits*. Retrieved from

Widrow, B., & Lehr, M. A. (1990). 30 years of adaptive neural networks: perceptron, madaline, and backpropagation. *Proceedings of the IEEE, 78*(9), 1415-1442.

Yuan, J., & Liberman, M. (2008). Speaker identification on the SCOTUS corpus. *Journal of the Acoustical Society of America, 123*(5), 3878.